%% file: main.tex
\documentclass[10pt,twocolumn,letterpaper]{article}

\usepackage[pagenumbers]{cvpr} 

\input{preamble}

\definecolor{cvprblue}{rgb}{0.21,0.49,0.74}
\usepackage[breaklinks,colorlinks,allcolors=cvprblue]{hyperref}
\hypersetup{
    pdftitle={RoboFailAtlas: Structured Reward Learning from Diverse Robot Failures},
    pdfauthor={Jiye Zhao, Kesen Zhao, Zehui Kong, Mu Wang, Fengda Zhang, Yin Li, Hanwang Zhang, Dongze Lian}
}
\usepackage{float}
\makeatletter
\patchcmd{\@maketitle}{\vskip .375in}{\vskip .15in}{}{}
\patchcmd{\@maketitle}{\vspace*{24pt}}{\vspace*{12pt}}{}{}
\patchcmd{\@maketitle}{\vspace*{12pt}\end{center}}{\vspace*{6pt}\end{center}}{}{}
\makeatother

\title{RoboFailAtlas: Structured Reward Learning from Diverse Robot Failures}

\author{
    Jiye Zhao$^{1,*}$ \quad
    Kesen Zhao$^{2,*}$ \quad
    Zehui Kong$^{1}$ \quad
    Mu Wang$^{1}$ \\
    Fengda Zhang$^{2}$ \quad
    Yin Li$^{3}$ \quad
    Hanwang Zhang$^{2}$ \quad
    Dongze Lian$^{1,\dagger}$ \\[0.35em]
    $^{1}$Nanjing University \qquad
    $^{2}$Nanyang Technological University \\
    $^{3}$Shanghai Xirang Kaiwu Technology Co., Ltd. \\[0.35em]
    {\small $^{*}$Equal contribution. \quad
    $^{\dagger}$Corresponding author.} \\[0.35em]
    {\small \textbf{Project Page: }
    \url{https://liangroup-nju.github.io/RoboFailAtlas-2026/}}
}
\begin{document}
\maketitle

\setlength{\stripsep}{0pt}

\begin{strip}
    \centering
    \includegraphics[width=0.96\linewidth]{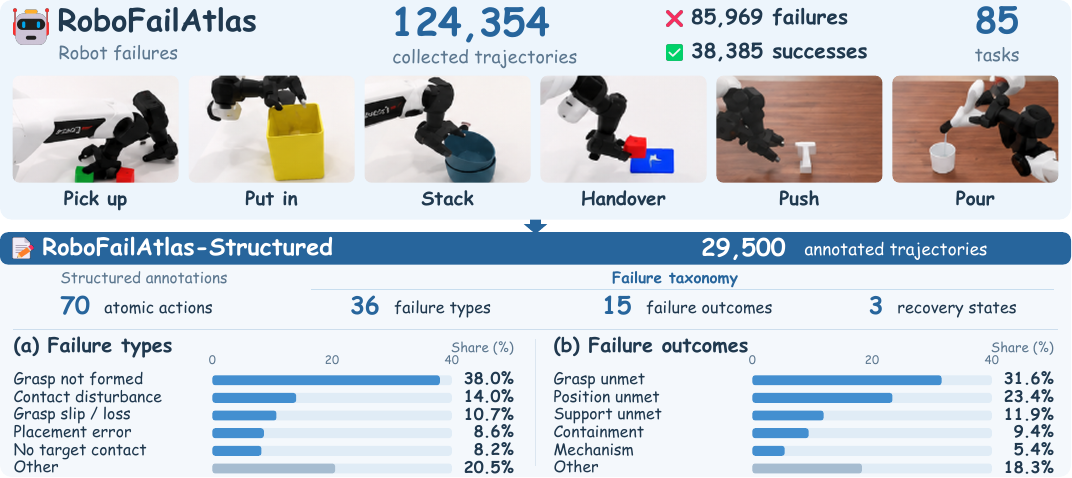}

    \captionsetup{hypcap=false}

    \captionof{figure}{
    Overview of our RoboFailAtlas and RoboFailAtlas-Structured.
    \textbf{Top:} 124,354 complete trajectories
    (85,969 failures and 38,385 successes) across
    85 tasks.
    \textbf{Bottom:} 29,500  trajectories with event-level structured annotations. The failure taxonomy includes 36 failure
    types, 15 failure outcomes, and three recovery states, while a vocabulary of 70 atomic actions captures execution context. Panels (a) and (b) show the distribution of failure types and outcomes, respectively.
    }

    \label{fig:teaser}
    \captionsetup{hypcap=true}
    \vspace{8pt}
\end{strip}

\input{sec/0_abstract}  
\vspace{-11pt}
\input{sec/1_intro}

\input{sec/2_related_work}
\input{sec/3_method}

\input{sec/4_experiment}

\input{sec/5_conclusion}

\section*{LLM Usage Statement}

OpenAI's ChatGPT was used to assist with manuscript
drafting, translation, language editing, and LaTeX
formatting. The authors reviewed and revised the
AI-assisted content.
Google Gemini-3.8-flash was used in the structured annotations pipeline to generate candidate annotations from robot execution trajectories under the proposed failure taxonomy. The annotations underwent automated
validation, model-assisted verification, and human
review, with flagged cases corrected as needed.
Qwen3-VL-4B-Instruct was used as the backbone for
structured reward learning, as described in the
methodology and experiments.
The authors take full responsibility for the
methodology, experimental results, and scientific
claims presented in this work.

{
    \small
    \setlength{\bibsep}{0pt plus 0.5pt}
    \bibliographystyle{ieeenat_fullname}
    \bibliography{main}
}

\input{sec/X_suppl}

\end{document}

%% file: preamble.tex
\usepackage{cuted}
\usepackage{capt-of}

%% file: sec/0_abstract.tex
\begin{abstract}

Recent work has suggested that failed trajectories capture a broader range of policy behaviors than successful demonstrations. However, large-scale failure data spanning diverse manipulation tasks with fine-grained, event-level annotations remain underexplored. To bridge this gap, we introduce RoboFailAtlas, a dataset containing 124,354 robot execution trajectories that capture diverse failures across 85 manipulation tasks. We define a failure taxonomy comprising 36 failure types, 15 failure outcomes, and three recovery states to characterize what goes wrong, which task conditions remain unmet, and whether the policy recovers. We also provide a subset of 29,500 trajectories with event-level structured annotations, which link each temporally localized failure event to its atomic-action segments and visual evidence while separately recording task progress and terminal success. We further propose a structured reward learning method that jointly predicts task completion and structured failure information. The resulting reward model outperforms existing baselines in progress estimation and terminal prediction while enabling fine-grained failure diagnosis. Across five tasks, PPO with structured rewards achieves an average success rate of 50.6\%, representing relative improvements of 26.5\% over task-only rewards and 24.0\% over environment rewards. Furthermore, failure-guided demonstration selection yields a 16.7\% relative improvement over random selection in simulation, while corrective demonstrations achieve an 8.9\% relative improvement over standard successful demonstrations on real robots, suggesting the potential of failure information to guide robot trajectory collection.

\end{abstract}

%% file: sec/1_intro.tex
\section{Introduction}
\label{sec:intro}

Successful demonstrations show how a robot completes a task but reveal little about the boundary of competent behavior~\cite{roboturk2018,dobbe2023,robomimic2022}. Failed executions complement them by exposing diverse deviations from success, including incorrect interactions, unmet prerequisites, repeated or ineffective actions, and recovery attempts across task stages~\cite{roboFAC,ViFailBack,reboot2026}. Failure trajectories therefore provide supervision beyond task completion, enabling models to characterize the boundary between successful and unsuccessful behavior, diagnose policy weaknesses, and learn rewards that reflect execution quality~\cite{aha2025,reflect2023,recoveryrl2021}.

Prior work has begun to transform unsuccessful executions into learning signals. Negative supervision can be constructed through counterfactual relabeling, temporal clipping, or targeted failure synthesis, while failed and suboptimal trajectories can be modeled using dense progress, execution-state labels, or trajectory comparisons~\cite{roboreward,densereward,robometer}. These approaches establish the value of learning from failures, but two aspects remain less explored. First, broad task coverage is rarely accompanied by diverse, fine-grained failures. Second, failure supervision often lacks an event-level representation that connects when a failure occurs with the action being executed, the task condition left unsatisfied, and the subsequent recovery. Without this structure, trajectories sharing the same terminal-success label may provide little information about how their execution processes differ.

Motivated by these gaps, we introduce \textbf{RoboFailAtlas},  as shown in Figure~\ref{fig:teaser}, a dataset of complete robot execution trajectories that captures diverse failures across a broad range of manipulation tasks. RoboFailAtlas contains 124,354 trajectories across 85 tasks, retaining successful and failed episodes, repeated attempts, and recovery behaviors. To complement this scale with fine-grained supervision, we construct RoboFailAtlas-Structured, an event-level annotated subset containing 29,500 trajectories across 50 tasks. We define a failure taxonomy along three dimensions: failure types, failure outcomes, and recovery states. Guided by this taxonomy, the structured annotations temporally localize failure events and ground them in atomic-action segments and visual evidence, while separately recording task progress and terminal success.

Building on these annotations, we develop a structured reward learning approach. A vision-language reward model jointly predicts task progress, terminal success, and structured failure information. We convert these predictions into a failure-aware reward that combines changes in task progress and terminal success with bounded penalties derived from predicted failure types and failure outcomes. This formulation distinguishes task completion from execution quality: progress and terminal success describe what the robot achieves, while structured failure predictions capture how its execution departs from successful behavior.

We evaluate our approach in reward modeling,
policy optimization, and failure-guided data selection.
Our reward model achieves a progress MAE of 0.0956,
a terminal ROC-AUC of 0.9949, and micro-F1 scores
above 0.93 for failure-associated actions, failure
types, and failure outcomes.
Across five tasks, the learned reward achieves
50.6\% average success, representing relative
improvements of 26.5\% over task-only rewards
and 24.0\% over environment terminal rewards.
Failure-guided demonstration selection yields
a 16.7\% relative improvement over random selection
in simulation, while corrective demonstrations
achieve an 8.9\% relative improvement over standard
successful demonstrations on real robots.
These results demonstrate the value of structured
failure supervision for reward and policy learning
and suggest its potential for guiding robot
data collection.

Our contributions are summarized as follows:

\begin{itemize}
    \item \textbf{A large-scale dataset of diverse robot failures.}
    We introduce RoboFailAtlas, comprising 124,354 complete trajectories across 85 manipulation tasks and covering successful executions, failures, retries, and recoveries.

    \item \textbf{Event-level structured annotations of robot failures.}
    We construct RoboFailAtlas-Structured, an annotated subset of 29,500 trajectories across 50 tasks, with temporally grounded failure events, a three-dimensional failure taxonomy, task progress, and terminal success.

    \item \textbf{Structured reward learning and policy improvement.}
    We develop a reward model that jointly predicts task completion and structured failure information, improving average policy success by 9.8 percentage points over environment terminal rewards across five tasks, and further explore failure-guided demonstration selection and corrective-data collection.

\end{itemize}

%% file: sec/2_related_work.tex
\section{Related Work}
\label{sec:related_work}

\paragraph{Robot Datasets.}
Large-scale robot datasets~\cite{
robonet2020,bridgedatav2_2023,openx2024,
droid2024,rh20t2023,agibotworld2025}
support policy learning across tasks and embodiments.
Simulation platforms~\cite{
robosuite2020,sapien2020,isaaclab2025,maniskill3_2024}
and manipulation benchmarks~\cite{
rlbench2020,libero2023,robocasa2024,robotwin2026}
enable scalable training and evaluation.
However, many datasets emphasize successful demonstrations,
underrepresenting errors, retries, and recoveries.
Prior work diversifies data through trajectory
generation~\cite{mimicgen2023},
expert feedback on policy-visited states~
\cite{dagger2011,hgdagger2019},
perturbed teleoperation~\cite{roboticdatainoculation2026},
and failure-aware selection~\cite{failuredata2026}.
Failure-inclusive resources include
RoboMIND~\cite{robomind2024} (erroneous executions);
DenseReward~\cite{densereward} (about 27,000 trajectories
from Isaac Sim failure synthesis and data from DROID,
robosuite, and LIBERO~\cite{droid2024,robosuite2020,libero2023});
and Robometer~\cite{robometer}
(successful, suboptimal, and failed trajectories).
We collect complete $\pi_{0.5}$~\cite{pi05_2025}
rollouts in RoboTwin and RoboDojo~
\cite{robotwin2026,robodojo2026},
preserving successes, failures, retries, and recoveries.

\begin{figure*}[!t]
    \centering
    \includegraphics[width=0.96\textwidth]{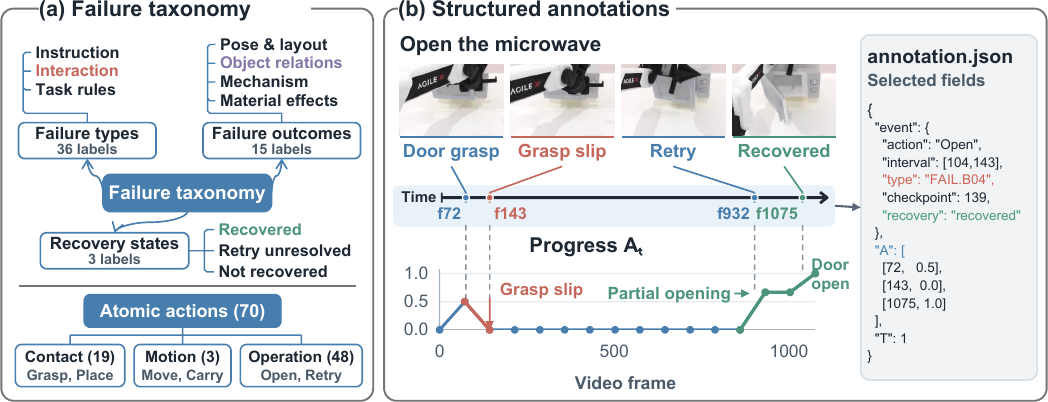}
    \caption{
    Overview of our failure taxonomy and structured annotations.
    \textbf{(a)} The taxonomy comprises 36 failure types,
    15 failure outcomes, and three recovery states,
    while 70 atomic actions provide execution context.
    \textbf{(b)} A microwave-opening trajectory illustrates
    grasp slippage, retry, and successful recovery,
    linking failure events and action segments to
    visual evidence and task-progress scores $A_t$.
    }
    \label{fig:structured_annotation}
    \vspace{-2pt}
\end{figure*}

\noindent\textbf{Robot Failures.}
Robot failure studies address detection, diagnosis,
and recovery.
Error-Aware Imitation Learning~\cite{errorawareil2021}
identifies potential failure states;
REFLECT~\cite{reflect2023} summarizes executions
for explanation and replanning;
AHA~\cite{aha2025} uses vision-language reasoning;
RoboFAC~\cite{roboFAC} organizes failures hierarchically;
and ViFailback~\cite{ViFailBack} provides visual diagnosis
and corrective guidance.
Recovery RL~\cite{recoveryrl2021} learns recovery policies;
REBOOT~\cite{reboot2026} provides phase-level annotations
and recovery demonstrations;
and WorldFailBench~\cite{worldFailBench} tests whether
world models reproduce policy failures.
Yet unified event-level annotations linking failures,
action context, unmet conditions, and recovery
remain underexplored.
Our taxonomy and structured annotations jointly capture
failure types, outcomes, and recovery states.

\noindent\textbf{Reward Learning.}
Robot reward learning draws on pretrained visual
representations (R3M~\cite{r3m2022}),
vision- and language-conditioned rewards
(VIP, LIV, and RoboCLIP~\cite{vip2023,liv2023,roboclip2023}),
preference learning~\cite{humanpreferences2017,pebble2021},
and vision-language model feedback~
\cite{vlmrm2024,rlvlmf2024}.
Failure-aware methods include
RoboReward~\cite{roboreward}
(negative and partial-progress examples),
DenseReward~\cite{densereward}
(dense progress rewards from synthesized failures),
and Robometer~\cite{robometer}
(progress supervision and trajectory preferences).
RoboValue~\cite{robovalue} evaluates progress tracking
and failure-recovery reasoning;
Robo-ValueRL~\cite{robovaluerl2026}
studies value estimation for policy learning.
Joint event-level modeling of failure types,
unmet task conditions, and recovery remains limited.
We train a reward model with task progress,
terminal outcomes, and structured failure supervision,
and evaluate failure-aware rewards in PPO.

%% file: sec/3_method.tex
\section{Method}
\label{sec:method}

Our framework comprises three stages: i) collecting successful and failed execution
trajectories and developing a failure taxonomy;
ii) constructing structured annotations that link temporally localized failures to atomic-action segments and visual evidence, while separately
recording task progress and terminal outcomes;
and iii) training a failure-aware reward model
for downstream policy optimization.

\subsection{Failure Taxonomy and Data Collection}
\label{sec:data-collection}

\paragraph{Failure Taxonomy.}
We initialize the failure taxonomy from task specifications~\cite{robotwin2026,robodojo2026}
and prior studies of robot execution failures~\cite{aha2025,densereward}.
We iteratively refine the taxonomy during data collection by reviewing observed failures
to identify missing, ambiguous, or overlapping categories.
Semantically equivalent failures share labels across tasks, while objects,
executing arms, and reference frames are represented as attributes.
Failures involving distinct evaluation criteria or state transitions
are assigned separate categories.

The resulting taxonomy comprises three complementary dimensions:
i) \textit{failure types} characterize execution anomalies, including instruction-following errors, physical-interaction failures, missing prerequisites, coordination errors, and violations of task constraints;
ii) \textit{failure outcomes} describe task conditions that remain unsatisfied at predefined evaluation checkpoints;
iii) \textit{recovery states} indicate whether a local failure is subsequently resolved, remains unresolved after another attempt, or is not followed by a recovery attempt.
The finalized taxonomy contains 36 failure types,
15 failure outcomes, and three recovery states.
Detailed definitions and category boundaries are provided in the supplementary material.

\begin{figure*}[!t]
    \centering
    \includegraphics[width=0.96\textwidth]{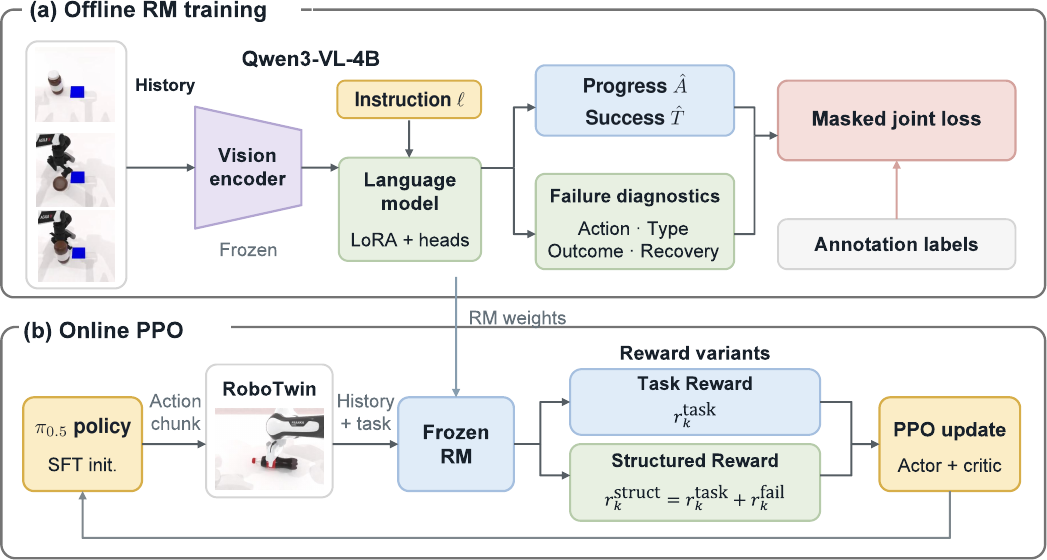}
    \caption{
    Structured Reward Learning and PPO.
    \textbf{(a)} Offline training of a Qwen3-VL-based
    reward model for task progress, terminal success,
    and structured failure prediction using a masked
    multitask loss.
    \textbf{(b)} Online PPO with frozen task-only
    and jointly supervised reward models for task
    and structured rewards, respectively.
    Structured rewards additionally incorporate
    failure feedback for policy optimization.
    }
    \label{fig:reward_model_ppo}
    \vspace{-2pt}
\end{figure*}

\noindent\textbf{Data Collection.}
We collect complete trajectories from autonomous
$\pi_{0.5}$~\cite{pi05_2025} executions in RoboTwin and
RoboDojo~\cite{robotwin2026,robodojo2026},
retaining both successful and failed episodes,
including execution errors, retries, and recovery behaviors.
RoboFailAtlas spans 85 manipulation tasks and contains
124,354 trajectories, with an approximate
failure-to-success ratio of 7:3.
We further construct RoboFailAtlas-Structured,
comprising 29,500 RoboTwin trajectories across
50 tasks with event-level structured annotations.
Detailed dataset statistics and splits are provided in the supplementary material.

\subsection{Structured Annotations}
\label{sec:annotation}

\paragraph{Atomic Actions and Failure Events.}
Following the taxonomy in Sec.~\ref{sec:data-collection},
we define 70 atomic-action categories and decompose
each trajectory into temporally localized segments.
Each segment records its action category, interval,
and task-specific attributes such as the executing
arm and manipulated object. Atomic actions provide
execution context rather than failure semantics.

We then define a failure event as the basic unit for instantiating the failure taxonomy on a trajectory. For a trajectory $\tau$, the $i$-th failure event is formalized as
\begin{equation}
    \begin{aligned}
    e_i = \bigl(
    &u_i, I_i,
    (\mathcal{S}_i^{\mathrm{ctx}},
     \mathcal{S}_i^{\mathrm{fail}}),
    &(F_i,O_i,R_i), E_i,U_i
    \bigr),
    \end{aligned}
    \label{eq:failure-event}
\end{equation}
where $u_i$ denotes the event identifier and $I_i$ its observed temporal interval. $\mathcal{S}_i^{\mathrm{ctx}}$ contains the associated atomic-action segments, while $\mathcal{S}_i^{\mathrm{fail}}\subseteq\mathcal{S}_i^{\mathrm{ctx}}$ contains only segments with evidence of local execution failure and may be empty. $F_i$ denotes the set of failure-type labels, $O_i$ records failure-outcome labels with their corresponding evaluation checkpoints, and $R_i$ indicates the recovery state. $E_i$ contains visual evidence and available action or state observations, while $U_i$ records field-level annotation status and uncertainty. A trajectory may contain multiple failure events and labels, with duplicates merged by object, attempt, checkpoint, and criterion. Uncertain fields are independently masked during training and evaluation.

\noindent\textbf{Step-Based Progress Scoring.}
Each task comprises $P$ phases with predefined
completion steps and dependencies, where phase
$p$ contains $N_p$ steps. At evaluation time $t$,
a multimodal model identifies completed, still-valid
steps $C_{p,t}$ in phase $p$ from observations,
and a program computes progress as
\begin{equation}
    A_t = \frac{1}{P}
    \sum_{p=1}^{P}
    \frac{|C_{p,t}|}{N_p}.
    \label{eq:progress-score}
\end{equation}
Only steps supported by observation evidence and satisfying their dependencies contribute to progress. Retries, waiting, and empty-hand movements do not directly increase $A_t$. Invalidated steps are removed from the completed set, allowing $A_t$ to decrease, while valid results after release or handover retain their contribution. Each trajectory has at most 16 progress evaluation points.

The terminal outcome $T$ is obtained from the simulator's binary success indicator and recorded independently of $A_t$. Although offline annotation may retrospectively inspect an entire trajectory, evidence used to assign a progress label at time $t$ must come from observations at or before $t$. During reward model inference, only the historical observations up to the current time are provided.

Finally, atomic-action segments, failure events, progress scores $A_t$, and terminal outcomes $T$ are organized into a unified JSON representation for reward model supervision. Figure~\ref{fig:structured_annotation} illustrates our failure taxonomy and structured annotations through a microwave-opening example involving grasp slippage, retry, and successful recovery.

\noindent\textbf{Annotation Generation and Quality Control.}
Our taxonomy-guided structured annotations pipeline
organizes task
instructions, videos, and action-state data, then
uses a multimodal model to generate candidate
annotations. Its prompt specifies frame-indexed
evidence, predefined progress steps and dependencies,
and a structured JSON schema, requiring independent,
evidence-grounded assessments of failure types,
outcomes, and recovery states with explicit uncertainty.
Additional details on the annotation guidelines and quality control are provided in the supplementary material.

Candidate annotations are refined through human spot checks
and iterative guideline updates, followed by automated
validation, model-assisted verification, and human review.
Flagged annotations undergo manual correction.

To assess annotation quality, two annotators independently
review a subset of trajectories, with a third reviewer
adjudicating discrepancies to establish the human reference.
Human--human agreement reaches 92.5\%, while
Gemini-generated annotations achieve 82.2\%
agreement with the human reference.
\subsection{Structured Reward Learning}
\label{sec:reward-model}

\paragraph{Joint Reward Model.}
Structured Reward Learning trains a multitask reward
model based on Qwen3-VL-4B-Instruct~\cite{qwen3vl2025}
using task and structured failure supervision
(Figure~\ref{fig:reward_model_ppo}(a)).
Given instruction $x$ and observation history
$\mathcal{H}_t$ up to time $t$, the model predicts
\begin{equation}
    f_\theta(x,\mathcal{H}_t)
    = (\hat A_t,\hat T_t,\hat{\mathbf z}_t),
    \label{eq:reward-model}
\end{equation}
where $\hat A_t$ and $\hat T_t$ denote predicted task progress and task-success scores, respectively. The structured prediction
$\hat{\mathbf z}_t =
(\hat{\mathbf z}_t^{\mathrm{act}},
\hat{\mathbf z}_t^{\mathrm{type}},
\hat{\mathbf z}_t^{\mathrm{out}},
\hat{\mathbf z}_t^{\mathrm{rec}})$
comprises failure-associated actions, failure types, failure outcomes, and recovery states.

The joint training objective combines task supervision
and structured failure supervision:
\begin{align}
    \mathcal{L}_{\mathrm{joint}}
    &= \mathcal{L}_{\mathrm{task}}
    + \lambda_{\mathrm{struct}}
      \mathcal{L}_{\mathrm{struct}},
    \label{eq:joint-loss}\\
    \mathcal{L}_{\mathrm{task}}
    &= \mathcal{L}_{\mathrm{prog}}(\hat A_t,A_t)
    + \mathcal{L}_{\mathrm{term}}(\hat T_t,T),
    \label{eq:task-loss}\\
    \mathcal{L}_{\mathrm{struct}}
    &= \sum_{j\in\mathcal{J}}
       \mathcal{L}_{j}
       (\hat{\mathbf z}_t^j,\mathbf z_t^j)
    + \lambda_{\mathrm{rec}}
      \mathcal{L}_{\mathrm{rec}}
      (\hat{\mathbf z}_t^{\mathrm{rec}},
       \mathbf z_t^{\mathrm{rec}}).
    \label{eq:struct-loss}
\end{align}

Here, $\mathcal{J}=\{\mathrm{act},\mathrm{type},
\mathrm{out}\}$ indexes the three multilabel
prediction heads, and $\mathbf z_t$ denotes
the structured reference labels.
$\mathcal{L}_{\mathrm{prog}}$ uses mean squared error,
while $\mathcal{L}_{\mathrm{term}}$ uses binary
cross-entropy. Each $\mathcal{L}_j$ uses multilabel
binary cross-entropy, and $\mathcal{L}_{\mathrm{rec}}$
uses multiclass cross-entropy.
The coefficients $\lambda_{\mathrm{struct}}$ and
$\lambda_{\mathrm{rec}}$ control the weights of
structured and recovery supervision, respectively. Terminal outcomes are supervised only at trajectory
endpoints. Missing or uncertain labels are masked,
and each loss is averaged over its valid supervision
entries. Detailed training configurations are provided
in the supplementary material.

\noindent\textbf{Policy Optimization with Structured Rewards.}
We integrate the trained reward model into PPO~\cite{ppo2017}, as illustrated in Figure~\ref{fig:reward_model_ppo}(b). After each action chunk, the frozen reward model $f_\theta$ evaluates the observation history and provides predictions for reward computation.

We construct the structured reward by combining task feedback and failure-related feedback:
\begin{equation}
    r_k^{\mathrm{struct}}
    =
    \underbrace{
        \hat A_k-\hat A_{k-1}+d_k\hat T_k
    }_{r_k^{\mathrm{task}}}
    +
    \underbrace{
        \bigl(-p_k^{\mathrm{type}}-d_kp_k^{\mathrm{out}}\bigr)
    }_{r_k^{\mathrm{fail}}}.
    \label{eq:structured-reward}
\end{equation}

Here, $k$ indexes action-chunk evaluation points, and $d_k$ indicates episode termination. The predictions $\hat A_k$ and $\hat T_k$ are obtained from the frozen reward model. The failure penalties $p_k^{\mathrm{type}}$ and $p_k^{\mathrm{out}}$ are derived from predicted failure types and outcomes, respectively, with confidence filtering, duplicate-penalty control, and magnitude limits. Detailed penalty construction is provided in the supplementary material. During PPO training, the reward model remains frozen, while the policy and value network are updated using the learned reward signals.

%% file: sec/4_experiment.tex
\section{Experiments}
\label{sec:experiment}

We evaluate RoboFailAtlas through five aspects:
reward prediction, downstream PPO optimization,
data-scale and history-length ablations,
failure-guided demonstration selection, and
real-world learning from failure corrections.

\subsection{Experimental Setup}
\label{sec:exp_setup}

\begin{table*}[!t]
    \centering
    \caption{
    Reward model evaluation on the RoboTwin test set.
    Task prediction is evaluated using progress MAE
    and terminal classification metrics.
    Structured prediction is evaluated using Micro-F1.
    }
    \label{tab:rm_benchmark}
    \small
    \setlength{\tabcolsep}{3pt}
    \renewcommand{\arraystretch}{1.15}

    \begin{tabular*}{\textwidth}{
        @{\extracolsep{\fill}}lcccccccc@{}}
        \toprule
        & \multicolumn{4}{c}{Task Prediction}
        & \multicolumn{4}{c}{Structured Prediction (Micro-F1)} \\
        \cmidrule(lr){2-5}
        \cmidrule(l){6-9}

        Model &
        \shortstack{$A$\\MAE $\downarrow$} &
        \shortstack{$T$\\AUC $\uparrow$} &
        \shortstack{Bal.\\Acc. $\uparrow$} &
        \shortstack{Success\\F1 $\uparrow$} &
        \shortstack{Action\\F1 $\uparrow$} &
        \shortstack{Type\\F1 $\uparrow$} &
        \shortstack{Outcome\\F1 $\uparrow$} &
        \shortstack{Recovery\\F1 $\uparrow$} \\
        \midrule

        Robometer-4B~\cite{robometer}
        & 0.3405 & 0.7483 & 71.13\% & 0.6016
        & -- & -- & -- & -- \\

        RoboReward-4B~\cite{roboreward}
        & 0.2536 & 0.5434 & 58.11\% & 0.5059
        & -- & -- & -- & -- \\

        RoboReward-8B~\cite{roboreward}
        & 0.2559 & 0.5183 & 50.63\% & 0.2020
        & -- & -- & -- & -- \\

        DenseReward~\cite{densereward}
        & 0.3355 & 0.6399 & 61.88\% & 0.5167
        & -- & -- & -- & -- \\

        \midrule

        Ours (Task-only)
        & \textbf{0.0857}
        & 0.9952
        & 97.72\%
        & 0.9673
        & -- & -- & -- & -- \\

        Ours (T-only, Full)
        & --
        & \textbf{0.9969}
        & \textbf{98.15\%}
        & \textbf{0.9683}
        & -- & -- & -- & -- \\

        Ours (Joint)
        & 0.0956
        & 0.9949
        & 97.58\%
        & 0.9636
        & \textbf{0.9376}
        & \textbf{0.9369}
        & \textbf{0.9682}
        & \textbf{0.7605} \\

        \bottomrule
    \end{tabular*}

    \vspace{3pt}
    \parbox{\textwidth}{%
        \footnotesize
        \emph{Note:} -- denotes unavailable metrics
        or predictions not evaluated under our taxonomy.
        T-only (Full) uses terminal-only supervision.
        Recovery predictions with maximum class probability
        below 0.5 are treated as abstentions.
        Additional structured metrics are provided
        in the supplementary material.
    }
    \vspace{-3pt}
\end{table*}

\begin{figure*}[t]
    \centering
    \includegraphics[width=0.96\textwidth]{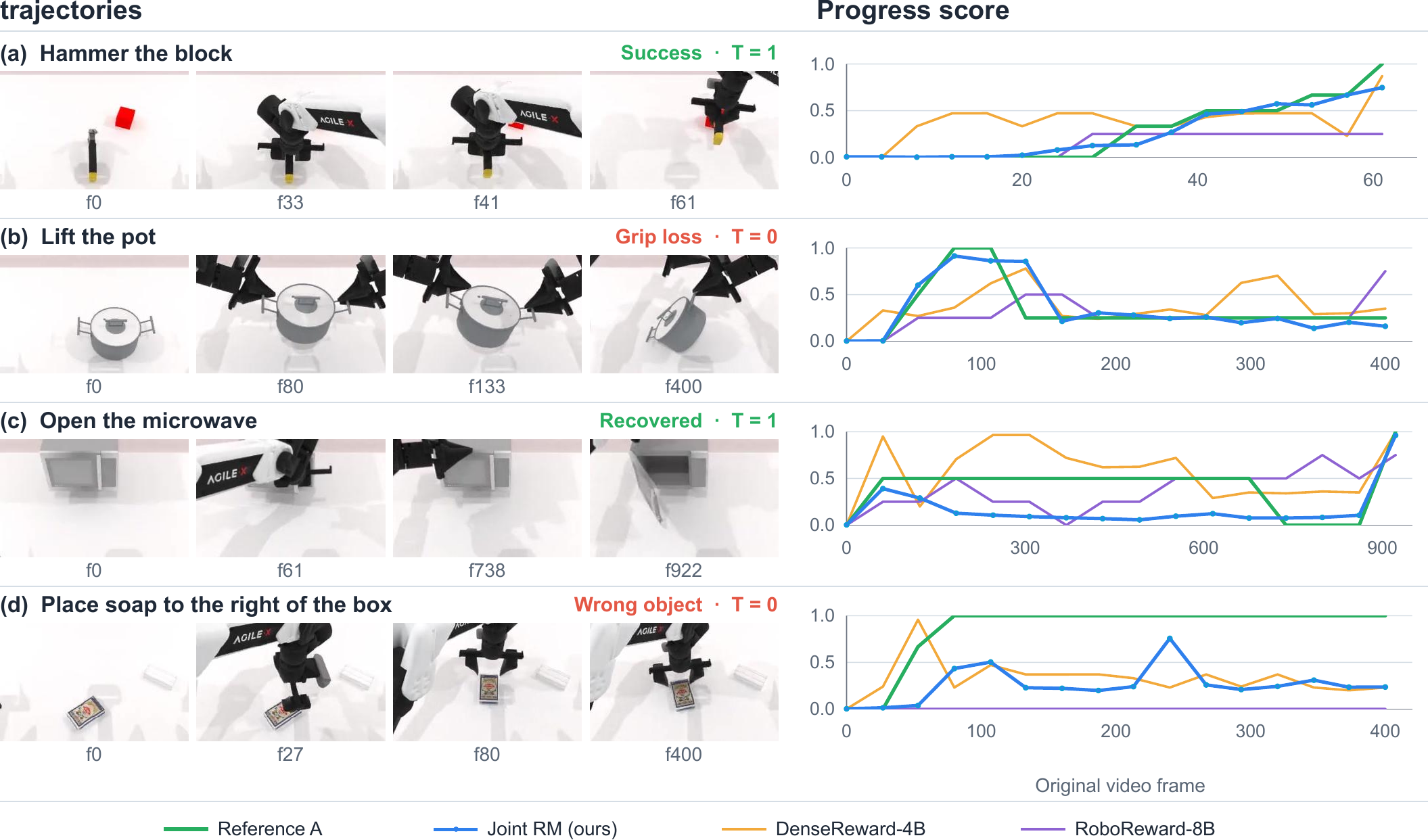}

    \caption{
    Qualitative task-progress predictions on RoboTwin
    trajectories. Left: sampled frames; right: progress
    scores over original video frame indices.
    \textbf{(a)} Successful block hammering ($T=1$).
    \textbf{(b)} Grip loss during pot lifting ($T=0$).
    \textbf{(c)} Microwave-opening failure and recovery ($T=1$).
    \textbf{(d)} Wrong-object manipulation during soap
    placement ($T=0$).
    Curves: reference $A_t$ (green), our joint RM (blue),
    DenseReward-4B (orange), and RoboReward-8B (purple).
    }
    \label{fig:failure_trajectory_progress}
    \vspace{-3pt}
\end{figure*}

\begin{figure*}[!t]
    \centering
    \includegraphics[width=0.99\textwidth]{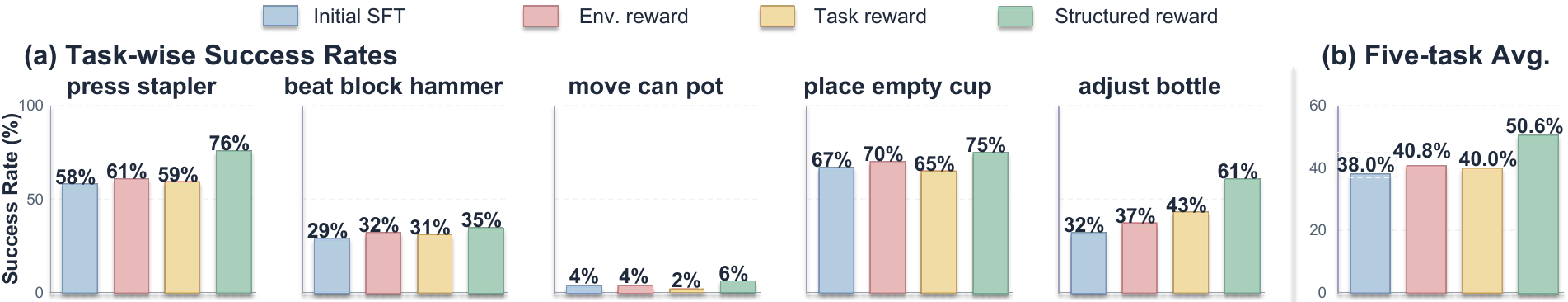}

    \caption{
    PPO policy success rates on five RoboTwin tasks.
    \textbf{(a)} Task-wise comparison of SFT,
    environment terminal rewards (Env.),
    task-only rewards (Task), and structured rewards
    (Struct.).
    \textbf{(b)} Average success rates:
    38.0\% (SFT), 40.8\% (Env.),
    40.0\% (Task), and 50.6\% (Struct.).
    }

    \label{fig:robotwin_reward_comparison}
    \vspace{-2pt}
\end{figure*}

\begin{figure*}[!t]
    \centering
    \includegraphics[width=0.98\textwidth]{
        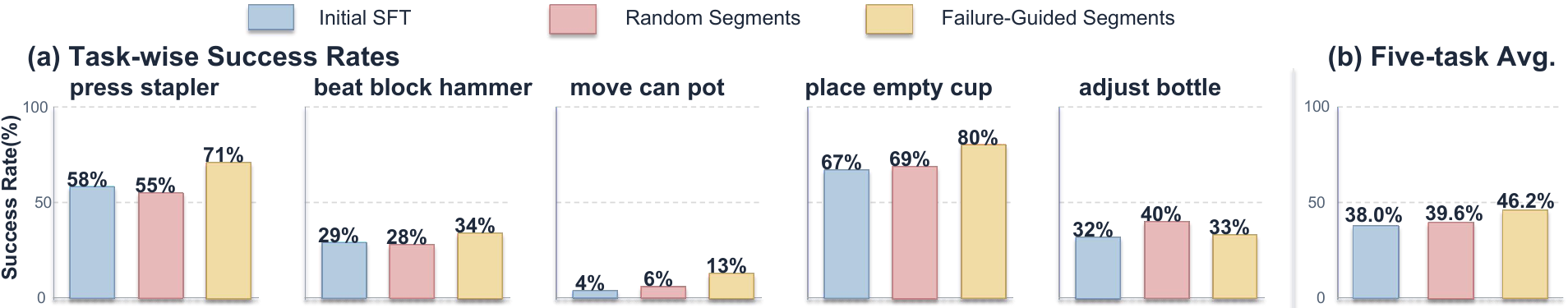
    }
    
    \caption{
    Failure-guided demonstration selection on five
    RoboTwin tasks.
    \textbf{(a)} Task-wise success rates of initial SFT,
    random selection, and failure-guided selection.
    \textbf{(b)} Average success rates: 38.0\%,
    39.6\%, and 46.2\%, respectively.
    }

    \label{fig:failure_guided_task_comparison}
    \vspace{-2pt}
\end{figure*}

\paragraph{Dataset and Splits.}
RoboFailAtlas-Structured contains 29,500 annotated trajectories across 50 RoboTwin tasks, split by trajectory into training, validation, and test sets (8:1:1). The T-only (Full) model additionally uses terminal-labeled training trajectories from RoboFailAtlas; all models share the same fixed RoboTwin test set.

\noindent\textbf{Reward Model Implementation.}
We use Qwen3-VL-4B-Instruct with task instructions and up to eight observation frames, including the current frame and uniformly sampled historical frames. We freeze the visual encoder and train LoRA adapters (rank 16, scaling 32) in the language module and prediction heads for four epochs (effective batch size 64).
The learning rates are $2\times10^{-5}$ for LoRA
parameters and $1\times10^{-4}$ for prediction heads.
For joint supervision, we set
$\lambda_{\mathrm{struct}}=0.1$ and
$\lambda_{\mathrm{rec}}=0.5$.
We train the Task-only and Joint models using
$\mathcal{L}_{\mathrm{task}}$ and
$\mathcal{L}_{\mathrm{joint}}$, respectively.
The T-only (Full) model uses only terminal-success
supervision with an expanded training set.
Missing or uncertain labels are excluded through masks.
Checkpoints are selected based on validation progress
MAE, with terminal ROC-AUC used to break ties.
Further implementation details are provided in
the supplementary material.

\noindent\textbf{Baselines and Metrics.}
We compare our reward models with publicly released
Robometer~\cite{robometer},
RoboReward~\cite{roboreward}, and
DenseReward~\cite{densereward} checkpoints.
Their native progress scores are mapped to $[0,1]$
using predefined rules, and endpoint scores serve
as proxies for terminal success.
All models are evaluated on the same test trajectories
and evaluation points. We evaluate task-progress estimation using mean
absolute error (MAE), and terminal success prediction
using ROC-AUC, balanced accuracy, and success-class F1.
For structured annotation prediction, we report
micro-precision, micro-recall, micro-F1, and macro-F1.
Classification thresholds are selected on the
validation set where applicable. Downstream policy
performance is evaluated using environment-defined
task success rates.

\subsection{Reward Model Evaluation}
\label{sec:rm_benchmark}

\paragraph{Progress Estimation and Terminal Prediction.}
Table~\ref{tab:rm_benchmark} shows that our task-only
and joint models outperform external baselines in
progress estimation and terminal prediction.
The Task-only model achieves the lowest progress
MAE (0.0857), while T-only (Full) achieves the
best terminal metrics (0.9969 AUC, 98.15\% balanced accuracy, and 0.9683 success F1).
The joint model maintains strong task prediction
while enabling structured failure diagnosis.

\noindent\textbf{Prediction of Structured Annotations.}
The joint model achieves micro-F1 scores of 0.9376,
0.9369, 0.9682, and 0.7605 for failure-associated
actions, failure types, failure outcomes, and
recovery states, respectively, enabling fine-grained
failure diagnosis alongside task prediction.
Figure~\ref{fig:failure_trajectory_progress}
shows that reference progress can decrease after
state invalidation and recover following successful
retries. Our joint model tracks these changes,
while some external models deviate from the
reference scores.

\subsection{Policy Optimization with Structured Rewards}
\label{sec:ppo}

We optimize $\pi_{0.5}$ with PPO using
RLinf~\cite{rlinfvla2026} on five Franka-embodiment
RoboTwin tasks. Three tasks come from the RLinf-VLA
benchmark, with \textnormal{adjust\_bottle} and
\textnormal{press\_stapler} added.
All variants start from the same task-specific
SFT policy and use either environment terminal
rewards, task rewards from a frozen task-only
reward model, or structured rewards from a
frozen jointly supervised reward model.
We also report SFT performance.
All methods are evaluated on the same 100
fixed scenes per task.

Figure~\ref{fig:robotwin_reward_comparison} shows
that structured rewards outperform environment
and task-only rewards on all five tasks,
achieving 50.6\% average success
(+9.8 and +10.6 percentage points, respectively).
The largest gains over environment rewards occur
on \textnormal{press\_stapler} (+15 points) and
\textnormal{adjust\_bottle} (+24 points),
while task-only rewards do not consistently
improve over environment rewards.

\subsection{Ablation and Efficiency Analysis}
\label{sec:ablation}

\paragraph{Training Task Coverage and Data Scale.}
We vary training coverage (10, 25, and 50 tasks)
under \emph{Fixed} and \emph{Scaled} data settings,
keeping other training and evaluation settings
unchanged. Table~\ref{tab:task_coverage} shows
that broader task coverage improves overall
prediction but can reduce performance on
shared tasks when data volume is fixed.
Increasing data scale further improves
prediction accuracy across tasks.

\begin{table}[t]
    \centering
    \caption{
    Effects of task coverage and training data scale
    on reward prediction.
    Fixed uses 4,739 trajectories, while Scaled
    increases data volume with task coverage.
    Shared 10 evaluates the same ten tasks.
    }
    \label{tab:task_coverage}
    \small
    \setlength{\tabcolsep}{2.5pt}
    \renewcommand{\arraystretch}{1.15}
    \begin{tabular}{@{}lrrrrr@{}}
        \toprule
        Setting & Tasks & Traj. &
        \shortstack{All 50\\$A$ MAE $\downarrow$} &
        \shortstack{All 50\\$T$ AUC $\uparrow$} &
        \shortstack{Shared 10\\$A$ MAE $\downarrow$} \\
        \midrule
        Fixed & 10 & 4,739
            & 0.1764 & 0.9749 & 0.1197 \\
        Fixed & 25 & 4,739
            & 0.1553 & 0.9824 & 0.1308 \\
        Fixed & 50 & 4,739
            & 0.1261 & 0.9910 & 0.1392 \\
        \midrule
        Scaled & 10 & 4,739
            & 0.1764 & 0.9749 & 0.1197 \\
        Scaled & 25 & 11,865
            & 0.1367 & 0.9861 & 0.1140 \\
        Scaled & 50 & 23,608
            & \textbf{0.0956}
            & \textbf{0.9949}
            & \textbf{0.1085} \\
        \bottomrule
    \end{tabular}
    \vspace{-2pt}
\end{table}

\noindent\textbf{Observation History and Inference Efficiency.}
Table~\ref{tab:history_length} shows that
longer histories improve reward prediction,
but gains diminish beyond 8 frames while
inference cost rises substantially.
We therefore use 8 frames for online PPO
to balance accuracy and efficiency.

\begin{table}[t]
    \centering
    \caption{
    Reward prediction versus input history length.
    B1/B4 denote total inference time (ms)
    for batch sizes 1/4.
    }
    \label{tab:history_length}
    \small
    \renewcommand{\arraystretch}{1.15}
    \begin{tabular*}{\columnwidth}{
        @{}r@{\extracolsep{\fill}}rrrr@{}}
        \toprule
        Frames &
        $A$ MAE $\downarrow$ &
        $T$ AUC $\uparrow$ &
        B1 (ms) $\downarrow$ &
        B4 (ms) $\downarrow$ \\
        \midrule
        4  & 0.0981 & 0.9923
           & \textbf{90.1} & \textbf{317.8} \\
        8  & 0.0956 & 0.9949
           & 147.5 & 569.0 \\
        12 & \textbf{0.0951} & \textbf{0.9969}
           & 215.2 & 927.6 \\
        \bottomrule
    \end{tabular*}
    \vspace{-3pt}
\end{table}

\subsection{Failure-Guided Demonstration Fine-Tuning}
\label{sec:failure_guided_finetuning}

On the five RoboTwin tasks in Sec.~\ref{sec:ppo},
we identify weak action categories from the
RoboFailAtlas-Structured training split and
select matching segments from successful
expert demonstrations. Random selection uses
the same demonstration pool, with matched
segment lengths and amounts of valid action
supervision. Both methods share the same
initial SFT policy, 1,000 optimizer updates,
and 100 fixed evaluation scenes per task.

Figure~\ref{fig:failure_guided_task_comparison}
shows 46.2\% average success, versus 39.6\%
for random selection and 38.0\% for initial SFT
(+6.6 and +8.2 percentage points).
Failure-guided selection outperforms random on
four of five tasks, gaining 16 and 11 points
on \textnormal{press\_stapler} and
\textnormal{place\_empty\_cup}, respectively,
but losing 7 points on
\textnormal{adjust\_bottle}, indicating
task-dependent benefits.

\subsection{Real-World Learning from Failure Corrections}
\label{sec:real_world}

\noindent\textbf{Experimental Setup.}
We evaluate corrective demonstrations on two
AgileX PiPER tasks~\cite{agilex_piper_manual_2025}:
\emph{Block Stacking} (stacking two blocks)
and \emph{Block Placement} (placing a block
in a small basket). Object positions vary
within predefined ranges; success requires
a stable target configuration for at least
three seconds after gripper release
and withdrawal.

\noindent\textbf{Data Collection and Training.}
Starting from the original pretrained
$\pi_{0.5}$ weights~\cite{pi05_2025},
we first perform SFT on shared successful
demonstrations $D_0$ to obtain $P_0$.
$P_{\mathrm{success}}$ and $P_{\mathrm{correct}}$
then continue SFT from $P_0$ using
$D_0\cup D_{\mathrm{success}}$ and
$D_0\cup D_{\mathrm{correct}}$, respectively.
$D_{\mathrm{success}}$ contains demonstrations
from normal initial states, whereas
$D_{\mathrm{correct}}$ contains operator
corrections that complete tasks from failure
states reached by $P_0$, including missed
grasps, object drops, and inaccurate placements.
Both methods use matched additional expert
action data and training budgets, with a
50/50 mixture of shared and additional
demonstrations.
\begin{figure}[t]
    \centering
    \includegraphics[width=0.9\columnwidth]{
        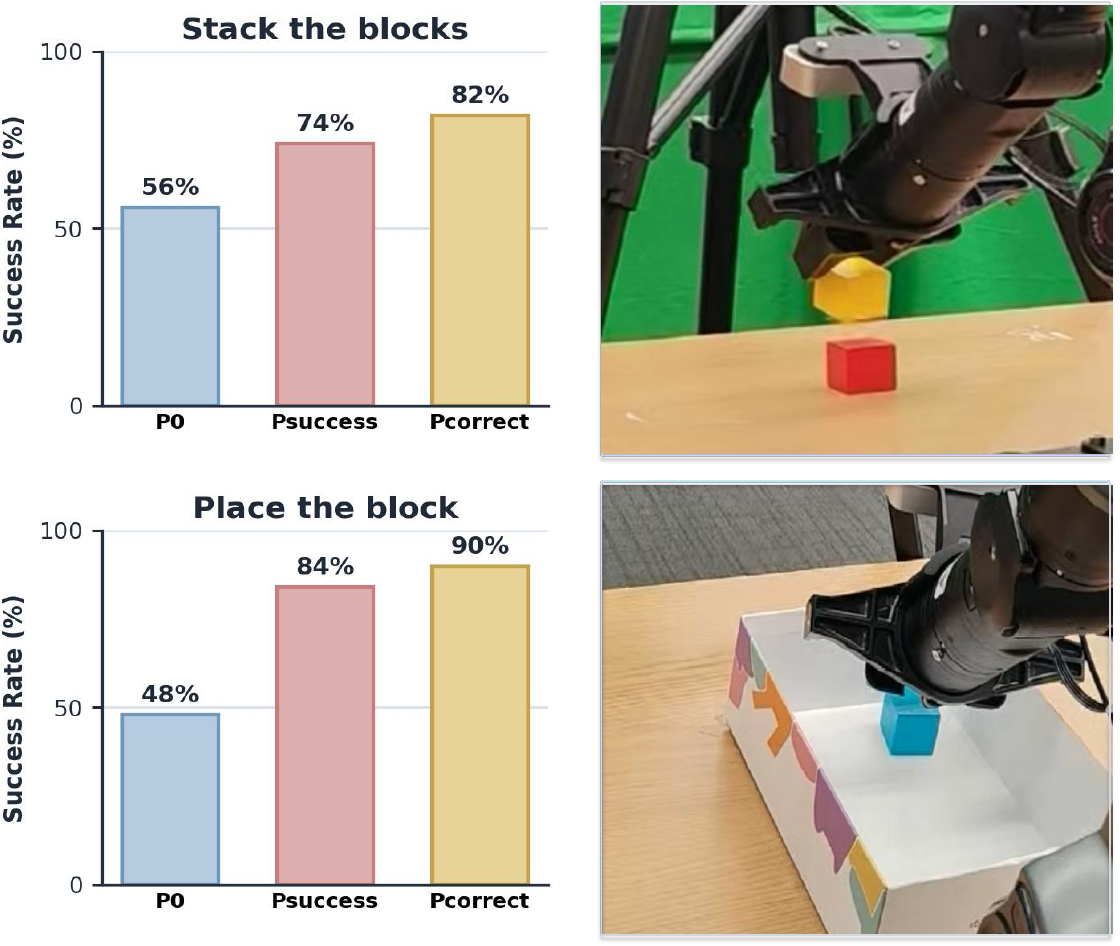
    }
    \caption{
    Real-world PiPER evaluation.
    \textbf{Top:} Block Stacking.
    \textbf{Bottom:} Block Placement.
    We compare frozen SFT ($P_0$), standard successful
    demonstration fine-tuning ($P_{\mathrm{success}}$),
    and failure-correction fine-tuning
    ($P_{\mathrm{correct}}$), with 50 trials per
    policy and task.
    }
    \label{fig:real_world_evaluation}
    \vspace{-3pt}
\end{figure}

\noindent\textbf{Evaluation Protocol.}
Each policy is evaluated over 50 autonomous
trials per task on matched test layouts.
As shown in Figure~\ref{fig:real_world_evaluation},
$P_{\mathrm{correct}}$ outperforms both
$P_0$ and $P_{\mathrm{success}}$ on both tasks,
improving average success by 7.0 percentage
points over $P_{\mathrm{success}}$.
These results demonstrate the benefit of
corrective demonstrations on real-world tasks.

%% file: sec/5_conclusion.tex
\section{Conclusion}
\label{sec:conclusion}
We introduce RoboFailAtlas, a dataset containing 124,354 robot execution trajectories that capture diverse failures across 85 manipulation tasks.
We define a failure taxonomy along three dimensions: failure types, failure outcomes, and recovery states, and use it to construct RoboFailAtlas-Structured, an event-level annotated subset of 29,500 trajectories. These annotations temporally localize failure events and ground them in atomic-action segments and visual evidence. Using this dataset, we train a reward model that jointly predicts task completion and structured failure information. The model outperforms existing baselines, while policies optimized with the resulting reward outperform those optimized with task-only or environment terminal rewards across five tasks. Exploratory studies further suggest the potential of failure information for demonstration selection in simulation and corrective data collection on real robots. These results demonstrate the value of RoboFailAtlas and the effectiveness of structured reward learning.

%% file: sec/X_suppl.tex
\clearpage
\appendix
\setcounter{section}{0}
\renewcommand{\thesection}{\Alph{section}}
\renewcommand{\thefigure}{\thesection.\arabic{figure}}
\renewcommand{\thetable}{\thesection.\arabic{table}}

\makeatletter
\@addtoreset{figure}{section}
\@addtoreset{table}{section}
\makeatother

\providecommand{\suppPending}{\textit{Pending}}

\providecommand{\suppFigure}[2]{%
  \IfFileExists{#1}{\includegraphics[width=\linewidth]{#1}}%
  {\fbox{\parbox{0.94\linewidth}{\centering Figure slot: #2}}}}

\providecommand{\suppFigurePlaceholder}[2]{%
  \fbox{\parbox[c][#1][c]{0.90\linewidth}{%
    \centering\small\textit{Figure placeholder}\par\vspace{0.7em}#2}}}

\raggedbottom
\setcounter{topnumber}{4}
\setcounter{bottomnumber}{3}
\setcounter{totalnumber}{7}
\setcounter{dbltopnumber}{3}

\renewcommand{\topfraction}{0.95}
\renewcommand{\dbltopfraction}{0.95}
\renewcommand{\textfraction}{0.05}
\renewcommand{\floatpagefraction}{0.75}
\renewcommand{\dblfloatpagefraction}{0.75}

\setlength{\textfloatsep}{8pt plus 1pt minus 1pt}
\setlength{\dbltextfloatsep}{8pt plus 1pt minus 1pt}
\setlength{\floatsep}{6pt plus 1pt minus 1pt}
\setlength{\dblfloatsep}{6pt plus 1pt minus 1pt}
\setlength{\parskip}{0pt}
\setlength{\tabcolsep}{4pt}
\renewcommand{\arraystretch}{1.0}

\captionsetup{
  font=small,
  labelfont=bf,
  skip=4pt,
  singlelinecheck=false
}

\makeatletter
\setlength{\@fptop}{0pt}
\setlength{\@dblfptop}{0pt}
\setlength{\@fpsep}{8pt}
\setlength{\@dblfpsep}{8pt}
\makeatother

\begin{center}
\textbf{Supplementary Material}
\end{center}

This appendix provides dataset statistics and annotation details
(Sections A--B), reward-model settings and loss-weight sensitivity (Section C),
policy-learning settings and additional PPO comparisons (Section D), and
qualitative trajectory examples (Section E).

\section{Dataset Composition and Filtering}
\label{app:dataset}

The complete collection contains 124,354 trajectories from RoboTwin and RoboDojo,
including both successful and failed policy executions
(Table~\ref{tab:supp_pool}). A task count refers to a platform--task pair.
Instructions, visual observations, environment outcomes, and available action/state
records are retained.

\begin{table}[!htbp]
\centering
\setlength{\tabcolsep}{3pt}
\small
\caption{Composition of the collected trajectory pool.}
\label{tab:supp_pool}
\begin{tabular}{lrrrr}
\toprule
Platform & Tasks & Successful & Failed & Total \\
\midrule
RoboTwin & 50 & 36,341 & 45,631 & 81,972 \\
RoboDojo & 35 & 2,044 & 40,338 & 42,382 \\
Total & 85 & 38,385 & 85,969 & 124,354 \\
\bottomrule
\end{tabular}
\end{table}

We initially select 600 trajectories per RoboTwin task, yielding 30,000 trajectories
across 50 tasks with a success-to-failure ratio of 3:7. Complete trajectories are
partitioned into training, validation, and test sets at an 8:1:1 ratio. Frames,
actions, and events from one trajectory remain in the same split.
Annotation-availability filtering retains 29,500 trajectories; the initial splits
lose 392, 59, and 49 trajectories, respectively
(Table~\ref{tab:robotiwn_subset}).

\begin{table}[!htbp]
\centering
\fontsize{8.5}{10.2}\selectfont
\setlength{\tabcolsep}{2pt}
\caption{Trajectory and scoring statistics for RoboFailAtlas-Structured.}
\label{tab:robotiwn_subset}
\begin{tabular}{lrrrr}
\toprule
Statistic & Training & Validation & Test & Total \\
\midrule
Successful & 7,189 & 899 & 900 & 8,988 \\
Failed & 16,419 & 2,042 & 2,051 & 20,512 \\
Trajectories & 23,608 & 2,941 & 2,951 & 29,500 \\
Records & 371,883 & 46,391 & 46,583 & 464,857 \\
Valid $A$ & 371,387 & 46,342 & 46,530 & 464,259 \\
Terminal $T$ & 23,608 & 2,941 & 2,951 & 29,500 \\
\bottomrule
\end{tabular}
\end{table}

Each trajectory has at most 16 candidate scoring timestamps, including its start
and endpoint. Individual uncertain fields are masked without discarding other known
labels. The test set contains 46,530 valid progress scores and 2,951 terminal labels.
The 20,512 environmentally failed trajectories contain 26,843 failure events.
A further 379 environmentally successful trajectories contain local failure events.

\paragraph{Full T-only Training Set.}
The expanded terminal-supervision experiment uses 118,354 training trajectories:
75,972 from RoboTwin and 42,382 from RoboDojo. All 6,000 originally reserved
RoboTwin validation/test identifiers are excluded from training.
Each training trajectory contributes one terminal example with environment-provided
$T$.

\begin{table}[!htbp]
\centering
\setlength{\tabcolsep}{3pt}
\small
\caption{Splits used by the Full T-only model.}
\label{tab:supp_full_split}
\begin{tabular}{lrrr}
\toprule
Split & RoboTwin & RoboDojo & Total \\
\midrule
Training & 75,972 & 42,382 & 118,354 \\
Validation & 2,941 & 0 & 2,941 \\
Test & 2,951 & 0 & 2,951 \\
\bottomrule
\end{tabular}
\end{table}

\section{Taxonomy and Annotation Method}
\label{app:taxonomy}

\paragraph{Sources of the Taxonomy.}
Our annotation design is informed by task specifications and prior failure-oriented
work, including AHA~\cite{aha2025} and DenseReward~\cite{densereward}. AHA studies
failure detection and natural-language reasoning; DenseReward uses synthesized
interaction failures for dense progress learning. We organize observed executions
into action context, execution anomalies, unmet task conditions, and recovery.

Descriptions are merged when they express the same observable failure criterion,
irrespective of object identity or task name. Labels are separated when they require
different evidence or describe different state transitions. For example, failing
to establish a grasp differs from losing an established grasp; a placement anomaly
differs from an unmet condition at a subsequent inspection. These dimensions can
coexist within one event.

\subsection{Atomic Actions}
\label{app:atomic_actions}

The vocabulary contains 70 atomic actions arranged into 13 functional groups and
three broad families: contact (19), motion (3), and operation (48). Action segments
record the category, executing arm, and inclusive start/end frame indices.
The annotation prompt represents a category using
\texttt{MOD.\textless{}action-group\textgreater{}/ATOM.\textless{}action-name\textgreater{}}; the reward-model head predicts the
70 \texttt{ATOM} categories. Actions describe what is executed: for example, a
missed hammering attempt can still be labeled \texttt{Hammer}.
The failure-associated action target is derived from the first confirmed action
execution failure.

\begin{table*}[!tp]
\centering
\begin{minipage}[t]{0.4773\textwidth}
\vspace{0pt}
\centering
\fontsize{8.5}{9.8}\selectfont
\setlength{\tabcolsep}{2pt}
\captionof{table}{Atomic-action vocabulary (1/2).}
\label{tab:supp_atoms_1}
\label{tab:atomic_actions}
\begin{tabular}{@{}p{0.24\linewidth}p{0.74\linewidth}@{}}
\toprule
Action & Functional group and definition \\
\midrule
Reach & \textit{acquire}. Approach an object before establishing a stable grasp. \\
Grasp & \textit{acquire}. Establish and maintain a stable grasp. \\
Pinch & \textit{acquire}. Grasp with fingertips or a small local contact area. \\
Lift & \textit{acquire}. Raise a contacted object from its supporting surface. \\
Move & \textit{transport}. Translate the end effector or object without specifying contact mode. \\
Carry & \textit{transport}. Transport an object while continuously holding it. \\
Takeout & \textit{transport}. Extract an object from a container or occluded region. \\
Place & \textit{release}. Set an object stably on a support or inside a container. \\
Release & \textit{release}. Remove a grasp without actively throwing the object. \\
Drop & \textit{release}. Release without continued support, allowing gravity-driven descent. \\
Throw & \textit{release}. Accelerate an object actively before releasing it. \\
Push & \textit{contact\_motion}. Apply contact force to move the target away from the contact side. \\
Pull & \textit{contact\_motion}. Draw the target toward the contact side. \\
Slide & \textit{contact\_motion}. Produce tangential motion while maintaining supporting contact. \\
PullApart & \textit{contact\_motion}. Separate connected or contacting objects using opposing forces. \\
Adjust & \textit{contact\_motion}. Make a local position or orientation correction. \\
Open & \textit{articulation}. Increase the opening of an articulated object. \\
Close & \textit{articulation}. Decrease the opening of an articulated object. \\
Rotate & \textit{rotation}. Change orientation around an axis. \\
Twist & \textit{rotation}. Apply torsional deformation or twisting contact. \\
Flip & \textit{rotation}. Turn an object over or reverse its supporting orientation. \\
Screw & \textit{rotation}. Engage or tighten a threaded connection. \\
Unscrew & \textit{rotation}. Disengage or loosen a threaded connection. \\
Insert & \textit{connection}. Move an object into a constrained opening or interface. \\
Connect & \textit{connection}. Establish a physical connection between objects. \\
Remove & \textit{connection}. Extract an object from a constraint or connection. \\
Press & \textit{control}. Apply normal pressure to a surface or control. \\
Toggle & \textit{control}. Switch a discrete control to a target setting. \\
Tap & \textit{control}. Make brief contact or a light impact. \\
Point & \textit{control}. Direct an end effector or finger toward a target. \\
Scan & \textit{control}. Move a scanner or target through the scanning region. \\
Pour & \textit{material}. Transfer liquid or particles by tilting a container. \\
Fill & \textit{material}. Add material continuously until the target amount is reached. \\
\bottomrule
\end{tabular}
\end{minipage}\hfill%
\begin{minipage}[t]{0.4773\textwidth}
\vspace{0pt}
\centering
\fontsize{8.5}{9.8}\selectfont
\setlength{\tabcolsep}{2pt}
\captionof{table}{Complete atomic-action vocabulary (2/2).}
\label{tab:supp_atoms_2}
\begin{tabular}{@{}p{0.24\linewidth}p{0.74\linewidth}@{}}
\toprule
Action & Functional group and definition \\
\midrule
Empty & \textit{material}. Discharge the contents of a container. \\
Scoop & \textit{material}. Collect material using a spoon, scoop, or shovel. \\
Stir & \textit{material}. Mix material through repeated circulating motion. \\
Shake & \textit{material}. Oscillate a held object back and forth. \\
Wipe & \textit{surface\_tool}. Slide a contacting tool across a surface to wipe it. \\
Brush & \textit{surface\_tool}. Move contacting bristles across a surface. \\
Sweep & \textit{surface\_tool}. Gather debris along a surface with a sweeping tool. \\
Hammer & \textit{surface\_tool}. Strike a target with a tool. \\
Peel & \textit{surface\_tool}. Progressively separate a thin layer from a substrate. \\
Fold & \textit{deformable}. Create a target fold in a deformable object. \\
Unfold & \textit{deformable}. Open an existing fold. \\
Stretch & \textit{deformable}. Increase a deformable object span through tension. \\
Straighten & \textit{deformable}. Straighten a curved or disordered deformable object. \\
Smooth & \textit{deformable}. Flatten wrinkles by moving across a deformable surface. \\
Hang & \textit{deformable}. Establish hanging support on a hook, rod, or hanger. \\
Untie & \textit{deformable}. Undo a knot. \\
Hold & \textit{handover\_hold}. Maintain possession when translation is not the primary objective. \\
HandOver & \textit{handover\_hold}. Transfer control of an object to another agent or arm. \\
TakeOver & \textit{handover\_hold}. Receive control of an object from another agent or arm. \\
Stabilize & \textit{handover\_hold}. Resist disturbances while supporting another operation. \\
ReturnToHome & \textit{control\_recovery}. Return the robot or end effector to its predefined home pose. \\
Retry & \textit{control\_recovery}. Repeat an operation after a failed attempt. \\
Recover & \textit{control\_recovery}. Restore an executable state after a failure or abnormal condition. \\
Tear & \textit{surface\_tool}. Split material by pulling it apart. \\
Cut & \textit{surface\_tool}. Sever material with a cutting edge. \\
Write & \textit{surface\_tool}. Produce a target trace using a writing tool. \\
Spray & \textit{material}. Discharge material through a nozzle. \\
Squeeze & \textit{contact\_motion}. Compress a deformable object, possibly expelling its contents. \\
Roll & \textit{contact\_motion}. Roll an object on a support or roll up deformable material. \\
Wrap & \textit{deformable}. Enclose an object with a flexible material. \\
Tie & \textit{deformable}. Form a knot in a cord or strap. \\
Zip & \textit{articulation}. Close a zipper. \\
Unzip & \textit{articulation}. Open a zipper. \\
Lock & \textit{articulation}. Engage a locking mechanism. \\
Unlock & \textit{articulation}. Disengage a locking mechanism. \\
Untangle & \textit{deformable}. Remove entanglement in cables or flexible objects. \\
Rinse & \textit{surface\_tool}. Wash an object surface with flowing water. \\
\bottomrule
\end{tabular}
\end{minipage}
\end{table*}

\subsection{Failure Types}
\label{app:failure_types}

Failure types describe observable execution anomalies. The 36 categories cover
instruction conformity (A01--A05), physical interactions (B01--B14), prerequisites
and coordination (D01--D10), and explicit task constraints (E01--E07).
Multiple labels are retained only when their criteria are independently supported.
We use the most specific applicable category and merge duplicate descriptions of
the same object, attempt, inspection, and criterion.

\begin{table*}[!tp]
\centering
\begin{minipage}[t]{0.4773\textwidth}
\vspace{0pt}
\centering
\fontsize{8.5}{10.2}\selectfont
\setlength{\tabcolsep}{2pt}
\captionof{table}{Failure-type taxonomy (1/2).}
\label{tab:supp_types_1}
\begin{tabular}{@{}p{0.19\linewidth}p{0.79\linewidth}@{}}
\toprule
Code & Failure type and observable criterion \\
\midrule
\texttt{FAIL.A01} & \textbf{Wrong manipulated object}. The operated object is outside the object or object set specified by the instruction. \\
\texttt{FAIL.A02} & \textbf{Wrong target or reference}. The operation is directed at a different receiving object, destination, or reference object. \\
\texttt{FAIL.A03} & \textbf{Wrong designated executor}. The arm or agent carrying out the operation differs from the explicitly designated executor. \\
\texttt{FAIL.A04} & \textbf{Wrong operation type}. The observed interaction is incompatible with the required operation and is not a valid preparation or alternative. \\
\texttt{FAIL.A05} & \textbf{Wrong state or direction}. The operation persistently acts toward a state or direction incompatible with the instruction. \\
\texttt{FAIL.B01} & \textbf{Target contact not established}. A contact attempt reaches its inspection point without establishing contact with the target. \\
\texttt{FAIL.B02} & \textbf{Wrong contact location or mode}. Contact occurs, but its location, tool end, or mode violates an explicit interaction requirement. \\
\texttt{FAIL.B03} & \textbf{Grasp not established after contact}. Target contact is observed, but the completed grasp attempt does not establish effective holding. \\
\texttt{FAIL.B04} & \textbf{Grasp slip or loss}. An established grasp undergoes unintended relative slip or loses the object. \\
\texttt{FAIL.B05} & \textbf{Required contact interrupted}. A required continuous non-grasp interaction contact is interrupted unexpectedly. \\
\texttt{FAIL.B06} & \textbf{Blocked or jammed motion}. An actual advancement attempt is impeded by an observable obstruction or constraint. \\
\texttt{FAIL.B07} & \textbf{Release does not free the object}. After a release attempt, the object remains physically constrained by the giving side. \\
\texttt{FAIL.B08} & \textbf{Placement pose incompatible with support}. The object is lowered or released with visibly incompatible support geometry and fails to seat normally. \\
\texttt{FAIL.B09} & \textbf{Placement position error}. An actual release or seating occurs outside the allowed placement region or relative-position range. \\
\texttt{FAIL.B10} & \textbf{Contact-induced disturbance}. Observed operation contact unexpectedly displaces, overturns, detaches, or reverses the state of an object or structure. \\
\texttt{FAIL.B11} & \textbf{Unexpected rebound}. An object unexpectedly bounces or recoils after contact, seating, or release, disrupting the required interaction. \\
\texttt{FAIL.B12} & \textbf{Non-contact alignment error}. The functional tool end, acting direction, or outlet is misaligned with the target during an observed operation. \\
\texttt{FAIL.B13} & \textbf{Rotation or adjustment magnitude error}. An executed rotation or adjustment undershoots or overshoots a verifiable target range at its inspection point. \\
\texttt{FAIL.B14} & \textbf{Hanging alignment error}. Hooking geometry is visibly misaligned during an actual hanging attempt, preventing engagement. \\
\bottomrule
\end{tabular}
\end{minipage}\hfill%
\begin{minipage}[t]{0.4773\textwidth}
\vspace{0pt}
\centering
\fontsize{8.5}{10.2}\selectfont
\setlength{\tabcolsep}{2pt}
\captionof{table}{Failure-type taxonomy (2/2).}
\label{tab:supp_types_2}
\begin{tabular}{@{}p{0.19\linewidth}p{0.79\linewidth}@{}}
\toprule
Code & Failure type and observable criterion \\
\midrule
\texttt{FAIL.D01} & \textbf{Advancement before prerequisite}. An operation begins before an explicitly required prerequisite state has been established. \\
\texttt{FAIL.D02} & \textbf{Premature handover release}. The giver actively releases before the receiver has established the required support or control. \\
\texttt{FAIL.D03} & \textbf{Auxiliary support removed too early}. An agent actively withdraws support or fixation while the dependent operation still requires it. \\
\texttt{FAIL.D04} & \textbf{Coordination relation violated}. Cooperating agents violate an explicit synchronization, relative-pose, or relative-motion requirement. \\
\texttt{FAIL.D05} & \textbf{Order or mutual-exclusion violation}. Observed operations violate an explicitly specified temporal order or mutual-exclusion window. \\
\texttt{FAIL.D06} & \textbf{Confirmed omission of required operation}. A necessary operation is demonstrably skipped despite a complete executable opportunity and no valid alternative. \\
\texttt{FAIL.D07} & \textbf{Local interaction absent or stalled}. A verifiable local opportunity shows sustained hovering, absent required interaction, or ineffective advancement. \\
\texttt{FAIL.D08} & \textbf{Gripper closure timing error}. The gripper closes before reaching a graspable approach position and continues toward the target while empty and closed. \\
\texttt{FAIL.D09} & \textbf{Release before support is established}. Outside direct handover, the holder actively releases before the required receiving support has been established. \\
\texttt{FAIL.D10} & \textbf{Tool action without the required tool}. The agent performs a tool-specific action without holding the required tool or an effective alternative. \\
\texttt{FAIL.E01} & \textbf{Prohibited contact or collision}. Contact explicitly prohibited by an applicable task rule occurs while that rule is active. \\
\texttt{FAIL.E02} & \textbf{Forbidden-region entry}. A robot or object enters a prohibited region or crosses an applicable spatial boundary. \\
\texttt{FAIL.E03} & \textbf{Protected object disturbed}. A protected non-target object changes position or state beyond the allowed range. \\
\texttt{FAIL.E04} & \textbf{Other invariant violated}. An explicit continuous-state requirement is broken and no more specific category expresses the same condition. \\
\texttt{FAIL.E05} & \textbf{Physical quantity exceeds a limit}. A task-defined physical quantity exceeds its threshold, supported by a reliable measured or calibrated signal. \\
\texttt{FAIL.E06} & \textbf{Task time or attempt budget violated}. An explicit deadline, duration window, attempt limit, or action-count budget is violated. \\
\texttt{FAIL.E07} & \textbf{Prohibited spillage}. Contents cross a specified boundary while an explicit no-spill rule is active. \\
\bottomrule
\end{tabular}
\end{minipage}
\end{table*}

\twocolumn[{
\begin{minipage}{\textwidth}

\subsection{Failure Outcomes}
\label{app:failure_outcomes}

Failure outcomes describe task conditions that remain unsatisfied at valid inspection points and comprise 15 categories. Outcome inspection can take place either when a local operation is completed or at the trajectory endpoint. A single event may have both a process-level failure type and an inspection-level outcome. For example, an observed positional deviation during placement can be annotated as \texttt{FAIL.B09}, while a subsequent inspection confirming that the target position has not been achieved can additionally be annotated as \texttt{FAIL.C01}.

\par\smallskip
\centering
\fontsize{8.5}{10.2}\selectfont
\setlength{\tabcolsep}{2pt}
\captionof{table}{Failure-outcome taxonomy.}
\label{tab:failure_outcomes}

\begin{minipage}[t]{0.4773\textwidth}
\vspace{0pt}
\begin{tabular}{@{}p{0.18\linewidth}p{0.80\linewidth}@{}}
\toprule
Code & Label and inspection criterion \\
\midrule
FAIL.C01 & \textbf{Position outcome not achieved}. The specified object's position, height, or distance from the target does not satisfy the requirement. \\
FAIL.C02 & \textbf{Orientation outcome not achieved}. The object's or tool's direction or angle does not satisfy the requirement. \\
FAIL.C03 & \textbf{Support relation not achieved}. The required support, stacking, or hanging relationship is not established. \\
FAIL.C04 & \textbf{Containment relation not achieved}. The specified object does not enter the target container or designated enclosed region. \\
FAIL.C05 & \textbf{Mechanism state not achieved}. The mechanism's rotation, travel, or position does not reach the required state. \\
FAIL.C06 & \textbf{Material transfer outcome not achieved}. The transferred quantity, proportion, distribution, or remaining amount does not meet the requirement. \\
FAIL.C07 & \textbf{Other explicit functional outcome not achieved}. A clearly specified functional effect is not produced, and no more specific outcome label applies. \\
FAIL.C08 & \textbf{Insertion relation not achieved}. The insertion depth, engagement, or mating relationship does not satisfy the requirement. \\
\bottomrule
\end{tabular}
\end{minipage}
\hfill
\begin{minipage}[t]{0.4773\textwidth}
\vspace{0pt}
\begin{tabular}{@{}p{0.18\linewidth}p{0.80\linewidth}@{}}
\toprule
Code & Label and inspection criterion \\
\midrule
FAIL.C09 & \textbf{Spatial arrangement not achieved}. The spatial order, spacing, or alignment among multiple objects does not satisfy the requirement. \\
FAIL.C10 & \textbf{Actuator signal endpoint not achieved}. The specified actuator control signal or measured state does not reach the required value. \\
FAIL.C11 & \textbf{Grasping relation not achieved}. The specified object and gripper or holder do not establish or maintain an effective grasp. \\
FAIL.C12 & \textbf{Handover control transfer not achieved}. The receiving side does not obtain control, or the giving side does not release its grasp. \\
FAIL.C13 & \textbf{Control-trigger outcome not achieved}. The button, lever, or other control does not produce the required triggering response. \\
FAIL.C14 & \textbf{Scanning or detection outcome not achieved}. The required scan coverage, detection completion, or effective detection signal is not achieved. \\
FAIL.C15 & \textbf{Dynamic-effect outcome not achieved}. The manipulation does not produce the required dynamic effect or observable motion. \\
\bottomrule
\end{tabular}
\end{minipage}

\end{minipage}
\vspace{8pt}
}]

\subsection{Recovery States and Auxiliary Annotations}
\label{app:recovery}

Recovery is evaluated against the same local condition as the original failure,
using evidence from the failure and a subsequent inspection. The model predicts
three known states; insufficient evidence is represented by \texttt{unknown} and
masked.

\par\smallskip
\noindent\begin{minipage}{\columnwidth}
\centering
\fontsize{8.5}{10.2}\selectfont
\setlength{\tabcolsep}{2pt}
\captionof{table}{Recovery states and annotation codes.}
\label{tab:recovery_states}
\begin{tabular}{@{}p{0.98\linewidth}@{}}
\toprule
Code and criterion \\
\midrule
\texttt{recovered}\newline The original local condition is restored at a later inspection. \\
\texttt{attempted\_not\_recovered}\newline A corrective attempt is observed, but the local condition remains unsatisfied. \\
\texttt{not\_recovered\_as\_of\_checkpoint}\newline The local condition remains unsatisfied at the cited inspection checkpoint. \\
\texttt{unknown}\newline Available evidence does not determine the recovery state. \\
\bottomrule
\end{tabular}
\end{minipage}
\par\smallskip

Executing-arm values are \texttt{left}, \texttt{right}, \texttt{both}, and
\texttt{unknown}. The action statuses \texttt{unknown} and \texttt{mixed} describe
uncertainty or an unresolved composite action. Cause analysis is recorded as
\texttt{hypothesis} or \texttt{unknown}, and consequences as \texttt{observed} or
\texttt{unknown}.
Task arm mode is \texttt{context.single}, \texttt{context.dual}, or \texttt{null}.
Successful trajectories may additionally be tagged \texttt{SUCCESS.REGULAR},
\texttt{SUCCESS.SUBOPTIMAL}, or \texttt{SUCCESS.UNASSESSED}. Suboptimal-success
features describe retries, recovery, redundant motion, unnecessary waiting,
or unstable adjustment, each linked to supporting observations.

\subsection{Process Completion $A$ and\texorpdfstring{\\}{ }Terminal Outcome~$T$}
\label{app:scoring}

For each task, a fixed template specifies phases, completion steps, and physical
dependencies. At time $t$, \texttt{gemini-3.8-flash} identifies completed steps that
remain valid, after which the program computes process completion:
\begin{equation}
A_t=\frac{1}{P}\sum_{p=1}^{P}\frac{|C_{p,t}|}{N_p}.
\label{eq:supp_progress}
\end{equation}
Here, $P$ is the number of phases, $N_p$ is the number of steps in phase $p$,
and $C_{p,t}$ is the set of completed steps that remain valid in that phase at time $t$.
Phases are equally weighted, and steps within each phase are equally weighted.
The template denominator is fixed. Repeating an attempt, waiting, moving an empty
gripper, or issuing a closure command does not independently increase progress.
An uncontrolled fall can invalidate previously completed operations; achievements
that remain valid after normal release continue to contribute. Unknown steps
produce an uncertain score, and the corresponding $A$ supervision and evaluation
are masked at that point.

For \texttt{handover\_block}, the template has four equally weighted phases:
giver control (\texttt{giver}), presentation (\texttt{present}), handover
(\texttt{receiver}, \texttt{transfer}), and placement (\texttt{placed}).
The steps follow giver control, presentation, receiver control, stable transfer,
and supported placement in that order. Completing the first two phases and one
of the two handover steps, with placement incomplete, gives
$A=(1+1+1/2+0)/4=0.625$.

The terminal label $T$ is the simulator's binary outcome and is stored separately
from $A$.

\subsection{Annotation Prompt Template}
\label{app:annotation_prompts}

The prompt template below summarizes the annotation instructions.
Task-specific evidence, taxonomy definitions, scoring steps, and a JSON Schema
are inserted for each trajectory.

\noindent\textbf{System Prompt.}
\begin{quote}\small\raggedright
Return only the final JSON object matching the requested schema. Do not include
analysis, reasoning, Markdown, or repeat the input. Use concise observations;
choose unknown when evidence is insufficient.
\end{quote}

\noindent\textbf{User Prompt (Template).}
\begin{quote}\small\raggedright
Annotate the robot trajectory using the task instruction, frame-indexed images,
available action/state evidence, and the supplied taxonomy.

\textbf{Inputs:}
\texttt{\{task\_instruction\_\allowbreak{}and\_platform\_outcome\}},
\texttt{\{trajectory\_evidence\}}, \texttt{\{visual\_evidence\_index\}},
\texttt{\{taxonomy\}}, \texttt{\{action\_scoring\_template\}},
and \texttt{\{response\_schema\}}.

\textbf{Actions.} Segment the observed execution into semantic actions, recording
category, arm, and inclusive start/end frames. Frame indices are zero-based;
references to the final segment array are one-based.

\textbf{Failures.} Record each confirmed event with its interval, inspection
frame, supporting observations, and associated actions. Distinguish contextual
segments from directly failed segments. Assess execution anomalies and unmet
inspection conditions independently, using only supported taxonomy codes.

\textbf{Recovery.} Evaluate the original local condition at a later inspection.
Choose \texttt{recovered}, \texttt{attempted\_not\_recovered},
\texttt{not\_recovered\_as\_of\_checkpoint}, or \texttt{unknown}.
A known state requires distinct images supporting the event and later inspection.

\textbf{Progress.} At each target frame, return completed and unknown step IDs,
evidence frames, and any rollback reason. Respect step dependencies and cite only
current or earlier observations. Repeated attempts do not themselves increase
progress; revoke achievements invalidated by a later state change.

\textbf{Evidence and output.} Use only supplied observations, preserve uncertain
fields as unknown, and do not infer causes from temporal order alone. Preserve
the environment outcome; a successful trajectory may contain local failures.
Return schema-valid JSON with \texttt{segments}, \texttt{context},
\texttt{events}, \texttt{first\_failure\_event}, \texttt{success},
\texttt{notes}, and \texttt{action\_scoring}.
\end{quote}

The program computes $A$ from step states and attaches environment outcome $T$
to the normalized export.

\subsection{Quality Control and Human Agreement}
\label{app:quality_control}

\begin{figure*}[!tp]
\centering
\includegraphics[width=\textwidth]{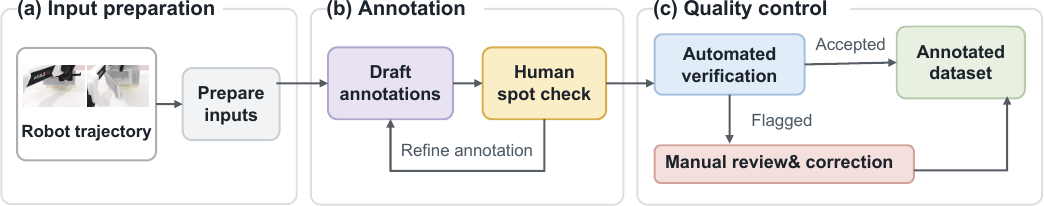}
\caption{Overview of structured trajectory annotation.
\textbf{(a) Input preparation:} Robot trajectories are organized into annotation
inputs.
\textbf{(b) Annotation:} Draft annotations are refined through human spot checks.
\textbf{(c) Quality control:} Automated verification retains accepted annotations
and flags questionable cases for manual review and correction before inclusion
in the annotated dataset.}
\label{fig:supp_annotation_workflow}
\end{figure*}

Figure~\ref{fig:supp_annotation_workflow} summarizes the annotation workflow.
Automated checks validate the schema, vocabulary, temporal ranges, segment
references, evidence frames, and consistency between step states and computed
progress. Questionable cases undergo further review and correction.
Two annotators independently review the same trajectories, and a third compares
their annotations and examines discrepancies to establish the human reference.
The recorded human--human agreement is \textbf{92.5\%}. Gemini-generated annotations
achieve \textbf{82.2\%} agreement with the finalized human reference.

The Gemini--human result is trajectory-averaged label-set Jaccard agreement:
\begin{equation}
\operatorname{Agreement}=\frac{1}{|\mathcal V|}
\sum_{i\in\mathcal V}\frac{|G_i\cap P_i|}{|G_i\cup P_i|}.
\label{eq:supp_agreement}
\end{equation}
$G_i$ and $P_i$ contain the unique failure-type and failure-outcome codes in the
reference and generated annotation. Only complete known references enter
$\mathcal V$. Two empty sets receive one; exactly one empty set receives zero.

Unknown reference fields are excluded independently through masks. Other known
fields in the same sample remain available for the corresponding training and
evaluation objectives.

\section{Reward Model Training and Evaluation}
\label{app:reward_model}

\subsection{Reward Model Settings}

\paragraph{Data Splits and Inputs.}
We use the trajectory-level training, validation, and test splits in
Section~\ref{app:dataset}; all scoring examples from the same trajectory remain
in the same split. Each example contains a task instruction and at most eight
distinct historical RGB frames, including the current frame. Earlier frames
are sampled uniformly from the causal prefix. Each frame is limited to 262,144
pixels. Simulator states and future observations are not model inputs.

\paragraph{Model and Supervision.}
We initialize the reward model from Qwen3-VL-4B-Instruct, freeze its vision
encoder, and train LoRA adapters in the language module and separate prediction
heads. The heads predict task progress, terminal success, failure-associated
actions, failure types, failure outcomes, and recovery states. Progress,
success, and multilabel outputs use sigmoid scores; recovery uses a three-class
softmax. Progress uses MSE, terminal and multilabel objectives use BCE, and
recovery uses multiclass cross-entropy. The Task-only model omits structured
losses and otherwise follows the corresponding common settings.

\begin{table}[!htbp]
\centering
\fontsize{8.5}{10.2}\selectfont
\setlength{\tabcolsep}{2pt}
\caption{Reward-model training configuration and checkpoint selection.}
\label{tab:reward_hyperparameters}
\begin{tabular}{@{}p{0.43\columnwidth}p{0.53\columnwidth}@{}}
\toprule
Parameter & Value \\
\midrule
Backbone & Qwen3-VL-4B-Instruct \\
Adaptation & Frozen vision encoder; language LoRA and prediction heads \\
LoRA rank / scaling & 16 / 32 \\
Precision & BF16 \\
RM training seed & 42 \\
Effective batch size & 64 \\
LoRA learning rate & $2\times10^{-5}$ \\
Head learning rate & $10^{-4}$ \\
Warmup ratio & 0.03 \\
Weight decay & 0.01 \\
Maximum pixels per image & 262,144 \\
Maximum sequence tokens & 8,192 \\
Historical frames, including current & At most 8 \\
Joint training & 4 epochs \\
$\lambda_{\rm struct}$ / $\lambda_{\rm rec}$ & 0.1 / 0.5 \\
Joint checkpoint & Lowest validation progress MAE; terminal AUC breaks ties; the same weights for task and diagnostic metrics \\
Task-only checkpoint & Lowest validation progress MAE; terminal AUC breaks ties \\
\bottomrule
\end{tabular}
\end{table}

\begin{table}[!htbp]
\centering
\setlength{\tabcolsep}{3pt}
\small
\caption{Terminal prediction of the selected Full T-only model on the fixed RoboTwin test set.}
\label{tab:supp_full_t_metrics}
\begin{tabular}{lrrr}
\toprule
Model & $T$ AUC $\uparrow$ & Bal. Acc. $\uparrow$ & Success F1 $\uparrow$ \\
\midrule
T-only (Full) & 0.9969 & 98.15\% & 0.9683 \\
\bottomrule
\end{tabular}
\end{table}

\subsection{Metrics and Evaluation Protocols}
\label{app:metrics}

Progress MAE uses known reference scores. Terminal prediction reports ROC-AUC,
balanced accuracy $(\operatorname{TPR}+\operatorname{TNR})/2$, and success-class F1.
Thresholds are selected on validation data and frozen for testing.
Multilabel micro-precision, recall, and F1 count TP, FP, and FN only on known
reference decisions. Macro-F1 averages over classes with at least one known
positive test example.
The diagnostic multilabel threshold is $0.8$. Recovery predictions with maximum
probability below $0.5$ abstain, contributing a false negative for the true class
without assigning a false positive to another class.

Unknown fields or categories are masked independently; other known labels in the
same sample remain eligible. At test time, 53 of 46,583 progress scoring points
are unavailable ($0.114\%$), while all 2,951 terminal outcomes are known.
Among the 2,951 test trajectories, at least one known reference category is
available for failure-associated actions, failure types, and failure outcomes in
2,700 ($91.49\%$), 2,795 ($94.71\%$), and 2,906 ($98.48\%$) trajectories,
respectively.

Recovery is known for 1,831 trajectories ($62.05\%$).

Table~\ref{tab:supp_structured_diagnostics} reports additional structured-diagnosis
metrics on the RoboTwin test set.

\begin{table}[!htbp]
\centering
\fontsize{8.5}{10.2}\selectfont
\setlength{\tabcolsep}{2pt}
\caption{Additional structured-diagnosis metrics on the RoboTwin test set.
Micro-F1 matches the corresponding main-paper results. All scores are fractions;
higher is better. Evaluation follows the known-reference masks defined above.}
\label{tab:supp_structured_diagnostics}
\begin{tabular}{lrrrr}
\toprule
Metric &
\shortstack{Failure-\\associated\\actions} &
\shortstack{Failure\\types} &
\shortstack{Failure\\outcomes} &
\shortstack{Recovery\\states} \\
\midrule
Classes & 70 & 36 & 15 & 3 \\
Micro-P $\uparrow$ & 0.8913 & 0.8924 & 0.9451 & 0.7749 \\
Micro-R $\uparrow$ & 0.9889 & 0.9861 & 0.9926 & 0.7466 \\
Micro-F1 $\uparrow$ & 0.9376 & 0.9369 & 0.9682 & 0.7605 \\
Macro-F1 $\uparrow$ & 0.7517 & 0.8077 & 0.9292 & 0.6691 \\
\bottomrule
\end{tabular}
\end{table}

\paragraph{Public Reward Models.}
\label{app:public_models}

Robometer and RoboReward use at most eight historical frames; DenseReward uses
its native three-frame input. RoboReward's ordinal score
$s\in\{1,\ldots,5\}$ is mapped to $(s-1)/4$. Native progress scores are used for
progress prediction and endpoint scores for terminal ranking.

\paragraph{Task Coverage and History Length.}
\label{app:task_coverage}
\label{app:history_latency}

The fixed-volume comparison uses 4,739 trajectories for 10, 25, and 50 tasks,
matching 74,556 scoring records. The scaled comparison uses all retained training
data for each task set. All groups use four fixed epochs, effective batch size 64,
seed 42, final checkpoints, and common frozen validation/test sets; the existing
10-task result is reused. The scaled setting changes both coverage and volume.
The history-length comparison uses causal prefixes of four, eight, or twelve
frames; B1/B4 denote total inference times for batches of one/four examples.

\subsection{Loss-Weight Sensitivity}
\label{app:loss_ablation}

We vary $\lambda_{\rm struct}\in\{0.05,0.10,0.20\}$ while fixing
$\lambda_{\rm rec}=0.5$, and compare $\lambda_{\rm rec}\in\{0.25,0.50,1.00\}$
at fixed $\lambda_{\rm struct}=0.1$. All groups use the complete structured training set
(23,608 trajectories, 50 tasks), seed 42, effective batch size 64, and the same
validation/test sets. The absolute recovery coefficient is
$\lambda_{\rm struct}\lambda_{\rm rec}$.

\begin{table*}[!tp]
\centering
\small
\setlength{\tabcolsep}{3pt}
\caption{Joint-loss weight sensitivity at four epochs.
Each nondefault row varies one weight around
$(\lambda_{\rm struct},\lambda_{\rm rec})=(0.10,0.50)$.
Diagnostic columns report endpoint Micro-F1;
higher is better except for progress MAE.}
\label{tab:supp_loss_weight_epoch4}
\begin{tabular}{rrrrrrrrrr}
\toprule
$\lambda_{\rm struct}$ &
$\lambda_{\rm rec}$ &
$A$ MAE $\downarrow$ &
$T$ AUC $\uparrow$ &
Bal. Acc. &
Success F1 &
Action &
Type &
Outcome &
Recovery \\
\midrule
0.05 & 0.50 & 0.0910 & 0.9944 &
\textbf{97.80\%} & \textbf{0.9705} &
0.9344 & 0.9332 & 0.9639 & 0.7512 \\

0.10 & 0.50 & 0.0956 & 0.9949 &
97.58\% & 0.9636 &
0.9376 & 0.9369 &
\textbf{0.9682} & \textbf{0.7605} \\

0.20 & 0.50 & 0.1024 & 0.9954 &
97.59\% & 0.9677 &
0.9366 & 0.9309 & 0.9652 & 0.7450 \\

0.10 & 0.25 & \textbf{0.0907} &
\textbf{0.9957} &
97.57\% & 0.9687 &
0.9322 & 0.9319 & 0.9639 & 0.7468 \\

0.10 & 1.00 & 0.1006 & 0.9956 &
97.51\% & 0.9620 &
\textbf{0.9442} & \textbf{0.9381} &
0.9681 & 0.7493 \\
\bottomrule
\end{tabular}
\end{table*}

Within the structural-weight sweep, a smaller weight gives lower progress MAE,
whereas a larger weight gives higher terminal AUC; the default structural weight
gives the highest Micro-F1 for all four structured outputs in this sweep.
Reducing the recovery weight to 0.25 improves progress MAE and terminal AUC,
whereas increasing it to 1.00 improves action and type Micro-F1. The default
configuration retains the highest outcome and recovery Micro-F1 among the
reported settings. These results show a trade-off between prediction objectives;
no single weight is best on every metric.

\section{Policy Learning and Full T-only PPO}
\label{app:policy}

\subsection{PPO Training}
\label{app:ppo}

\paragraph{Initialization and Implementation.}
We use RLinf's PPO implementation with RoboTwin's Franka configuration.
All five tasks start from the same $\pi_{0.5}$ SFT checkpoint at update 40,000.
The policy generates action chunks, and PPO updates the actor and value network.
When learned rewards are used, the reward model remains frozen.
Table~\ref{tab:supp_ppo_hyperparameters} lists the common settings;
Table~\ref{tab:supp_ppo_task_settings} lists the verified evaluation checkpoints
and task-specific settings of the Structured experiments.

\par\smallskip
\noindent\begin{minipage}{\columnwidth}
\centering
\fontsize{8.5}{10.2}\selectfont
\setlength{\tabcolsep}{2pt}
\captionof{table}{Common PPO configuration.}
\label{tab:supp_ppo_hyperparameters}
\begin{tabular}{@{}p{0.57\columnwidth}p{0.39\columnwidth}@{}}
\toprule
Parameter & Value \\
\midrule
Parallel training environments & 24 \\
Rollout collection epochs per iteration & 1 \\
Maximum control steps per environment per rollout & 400 \\
Interaction-step budget cap per iteration & $24\times400=9{,}600$ \\
Actions per chunk & 50 \\
Effective / micro-batch size & 64 / 2 \\
PPO optimization epochs per rollout batch & 2 \\
Policy / value learning rate & $5\times10^{-6}$ / $10^{-4}$ \\
PPO clip ratio & 0.2 \\
Discount factor $\gamma$ / GAE coefficient $\lambda$ & 0.99 / 0.95 \\
Entropy coefficient & 0 \\
Policy training seed & 0 \\
\bottomrule
\end{tabular}
\end{minipage}
\par\smallskip

\begin{table*}[!tp]
\centering
\fontsize{8.5}{10.2}\selectfont
\setlength{\tabcolsep}{5pt}
\caption{Training budgets and task-specific settings for the Structured
five-task experiments. A dash denotes an option not explicitly set in the
corresponding run configuration.}
\label{tab:supp_ppo_task_settings}
\begin{tabular}{@{}lrrrrrr@{}}
\toprule
Task &
\shortstack{Maximum\\steps} &
\shortstack{PPO\\iterations} &
\shortstack{Interaction\\budget cap} &
\shortstack{Critic-only\\warmup} &
\shortstack{SFT KL\\coefficient} &
\shortstack{Rollout KL\\target per chunk} \\
\midrule
\texttt{press\_stapler} & 400 & 50 & 480,000 & 10 & 0.1 & 0.02 \\
\texttt{beat\_block\_hammer} & 400 & 20 & 192,000 & 10 & 0.1 & 0.02 \\
\texttt{move\_can\_pot} & 400 & 20 & 192,000 & 10 & 0.1 & 0.02 \\
\texttt{place\_empty\_cup} & 400 & 20 & 192,000 & 10 & 0.1 & 0.02 \\
\texttt{adjust\_bottle} & 400 & 100 & 960,000 & -- & 0 & -- \\
\bottomrule
\end{tabular}
\end{table*}

\paragraph{Reward Construction.}
Environment reward is the terminal binary simulator success indicator.
The learned $A/T$ reward and structured reward follow the main-paper definitions:
\begin{align}
r_k^{A/T}&=\hat A_k-\hat A_{k-1}+d_k\hat T_k,\\
r_k^{\rm struct}&=r_k^{A/T}-p_k^{\rm type}-d_kp_k^{\rm out},
\end{align}
where $k$ indexes action-chunk scoring steps and $d_k$ indicates termination.
Both the progress increment and terminal score have unit coefficients.
Confidence filtering, repeated-penalty control, and magnitude caps are described
in Section~\ref{app:structured_rewards}. The Full T-only reward is defined in
Section~\ref{app:full_t_ppo}.

\paragraph{Evaluation and Checkpoints.}
Each policy uses a fixed PPO checkpoint and is evaluated on its task's cohort
of 100 fixed scenes. Success rates use the simulator's binary success indicator,
and the five-task average weights tasks equally.

\subsection{Structured Rewards}
\label{app:structured_rewards}

Categories are calibrated on 500 validation endpoints, ten per task, using stable
identifier hashing and task round-robin selection.

\begin{table}[!htbp]
\centering
\setlength{\tabcolsep}{3pt}
\small
\caption{Structured-reward parameters.}
\label{tab:structured_reward_config}
\begin{tabular}{p{0.54\columnwidth}p{0.36\columnwidth}}
\toprule
Parameter & Value \\
\midrule
High-confidence threshold & 0.8 \\
Low-confidence threshold & 0.5 \\
Consecutive confirmation chunks & 2 \\
Failure-type cost scale & 0.025 \\
Per-chunk type-cost cap & 0.05 \\
Per-trajectory type-cost cap & 0.10 \\
Terminal outcome-cost scale/cap & 0.05 \\
Terminal success gate & $\hat T<0.8$ \\
\bottomrule
\end{tabular}
\end{table}

A type is confirmed when both its score and the maximum score among its supported
actions reach 0.8 for two consecutive chunks. The first confirmation incurs
0.025 times its score, subject to per-chunk and per-episode caps. Each type is
penalized at most once per episode.

At termination, an outcome cost applies only if $\hat T<0.8$ and an enabled
outcome reaches 0.8. For the maximum qualifying score $p_{\max}$,
\begin{equation}
p_k^{\rm out}=0.05p_{\max}(1-\hat T_k).
\label{eq:outcome_cost}
\end{equation}
Otherwise the cost is zero.

\subsection{Failure-Label-Guided SFT}
\label{app:failure_guided_sft}

The five tasks match the PPO study. Failure annotations from the training split
identify error-prone action stages; task-specific gripper and kinematic rules
locate corresponding stages in successful expert demonstrations. Guided and
matched-random groups use the same demonstration pool, matching segment lengths
and valid action supervision. Full context is retained while masks restrict
action-loss targets.

Both groups start at SFT update 40,000 and train for 1,000 updates with effective
batch size 64 and seed 42. The learning rate decays from $5\times10^{-6}$ to
$10^{-6}$ after 50 warmup updates. Evaluation uses 100 fixed scenes per task.

\newpage
\subsection{PPO with Full T-only Rewards}
\label{app:full_t_ppo}

This experiment uses the Full T-only model from Section~\ref{app:reward_model}
with the common policy initialization and PPO settings. Its learned reward is
\begin{equation}
r_k^{\rm FullT}=d_k\hat T_k.
\end{equation}
The continuous sigmoid score is used to evaluate terminal histories.
Evaluation uses 100 fixed native scenes per task.

\par\smallskip
\noindent\begin{minipage}{\columnwidth}
\centering
\setlength{\tabcolsep}{3pt}
\small
\setlength{\tabcolsep}{5pt}
\captionof{table}{PPO with Full T-only rewards. Success rates use 100 fixed scenes per task, and the average weights the five tasks equally.}
\label{tab:supp_full_t_ppo}
\begin{tabular}{lrr}
\toprule
Task & Test scenes & Full T-only \\
\midrule
press\_stapler & 100 & 62.0\% \\
beat\_block\_hammer & 100 & 30.0\% \\
move\_can\_pot & 100 & 7.0\% \\
place\_empty\_cup & 100 & 65.0\% \\
adjust\_bottle & 100 & 29.0\% \\
Equal-task average & 500 & 38.6\% \\
\bottomrule
\end{tabular}
\end{minipage}
\par\smallskip

\begin{figure*}[!t]
\section{Qualitative Trajectories}
\label{app:qualitative}

\begin{minipage}{\textwidth}

Figure~\ref{fig:qualitative_trajectories} presents three RoboTwin annotation
examples illustrating partial completion, environment
success without full visual confirmation, and rollback of previously completed
operations. The curves show reference $A$.

\par\smallskip
\centering

\includegraphics[width=\linewidth]{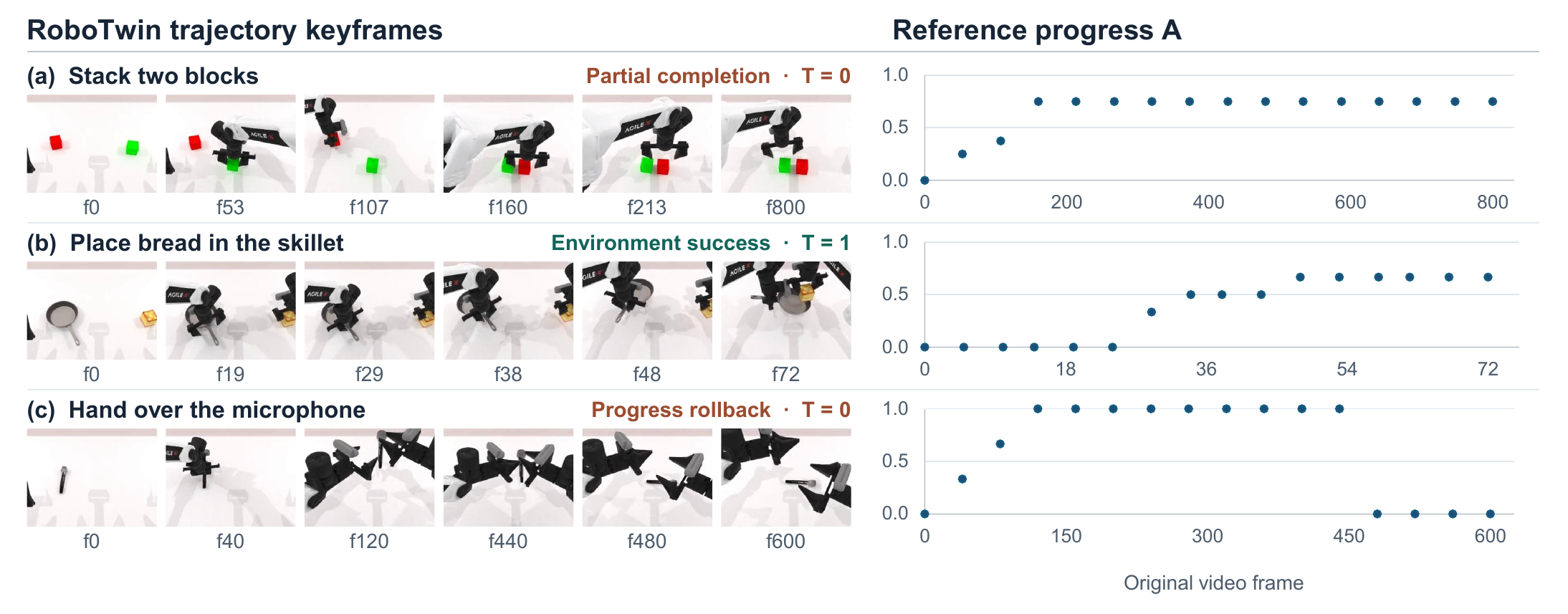}

\captionof{figure}{Representative trajectories with reference progress $A$
and terminal outcome $T$.
(a) Two blocks are transported and placed side by side; stacking remains
incomplete and endpoint $A=0.75$, $T=0$.
(b) The environment reports success while the bread remains held in the
available final view; release and support by the pan are not yet visually
confirmed, giving endpoint $A=2/3$, $T=1$.
(c) The microphone falls after handover; reference progress decreases from
one at frame 440 to zero at frame 480, and terminal $T=0$.}
\label{fig:qualitative_trajectories}

\end{minipage}
\vspace{8pt}
\end{figure*}